\pdfoutput=1
\documentclass[runningheads]{llncs}
\usepackage{graphicx}

\usepackage{tikz}
\usepackage{comment}
\usepackage{amsmath,amssymb} 
\usepackage{color}

\usepackage[accsupp]{axessibility}  

\usepackage{booktabs}
\usepackage{epsfig}
\usepackage{physics}
\usepackage{caption}
\usepackage{subcaption}
\usepackage{multirow}
\usepackage{array}
\usepackage{gensymb}
\usepackage{enumerate}
\usepackage{algorithm}
\usepackage{algorithmic}
\usepackage{makecell}
\usepackage{hyperref}

\DeclareMathOperator*{\argmax}{arg\,max}

\DeclareMathOperator*{\avg}{avg} 

\newcommand\blfootnote[1]{%
  \begingroup
  \renewcommand\thefootnote{}\footnote{#1}%
  \addtocounter{footnote}{-1}%
  \endgroup
}

\newcommand{\tabincell}[2]{\begin{tabular}{@{}#1@{}}#2\end{tabular}}
\usepackage{color}

\newcommand{\secb}[1]{{\color{red}\textbf{#1}}\normalfont}
\newcommand{\thrb}[1]{{\color{blue}\textbf{#1}}\normalfont}

\begin{document}
\pagestyle{headings}
\mainmatter
\def\ECCVSubNumber{2707}  

\title{DISP6D: Disentangled Implicit Shape and Pose Learning for Scalable 6D Pose Estimation} 

\titlerunning{DISP6D}
%
\author{Yilin Wen\inst{1} 
\and Xiangyu Li\inst{2}
\and Hao Pan\inst{3}
\and Lei Yang\inst{1,4}
\and Zheng Wang\inst{5}
\and Taku Komura\inst{1}
\and Wenping Wang\inst{6}
}
\authorrunning{Y. Wen et al.}
%
\institute{The University of Hong Kong \and Brown University \and Microsoft Research Asia \and Centre for Garment Production Limited, Hong Kong \and SUSTech \and Texas A\&M University}
\maketitle

\begin{abstract}

Scalable 6D pose estimation for rigid objects from RGB images aims at handling multiple objects and generalizing to novel objects. Building on a well-known auto-encoding framework to cope with object symmetry and the lack of labeled training data, we achieve scalability by disentangling the latent representation of auto-encoder into shape and pose sub-spaces. The latent shape space models the similarity of different objects through contrastive metric learning, and the latent pose code is compared with canonical rotations for rotation retrieval. Because different object symmetries induce inconsistent latent pose spaces, we re-entangle the shape representation with canonical rotations to generate shape-dependent pose codebooks for rotation retrieval. We show state-of-the-art performance on two benchmarks containing textureless CAD objects without category and daily objects with categories respectively, and further demonstrate improved scalability by extending to a more challenging setting of daily objects across categories.
\blfootnote{Work partially done during internships of Y. Wen and X. Li with Microsoft Research Asia. Code and data are available at: \url{https://github.com/fylwen/DISP-6D}.}

\end{abstract}


\section{Introduction}

Estimating the 6D pose of objects from a single RGB image is fundamental in fields like robotics and scene understanding.
While efficient learning-based methods have been developed \cite{rad2017bb8,xiang2017posecnn,kehl2017ssd}, a common assumption with many of these works is that a specialized network is trained for each object, which makes it expensive to process multiple objects by switching and streaming to respective networks, and renders it impossible to handle novel objects without re-training.

\begin{figure}[t]
	\centering
	\includegraphics[width=1.\linewidth]{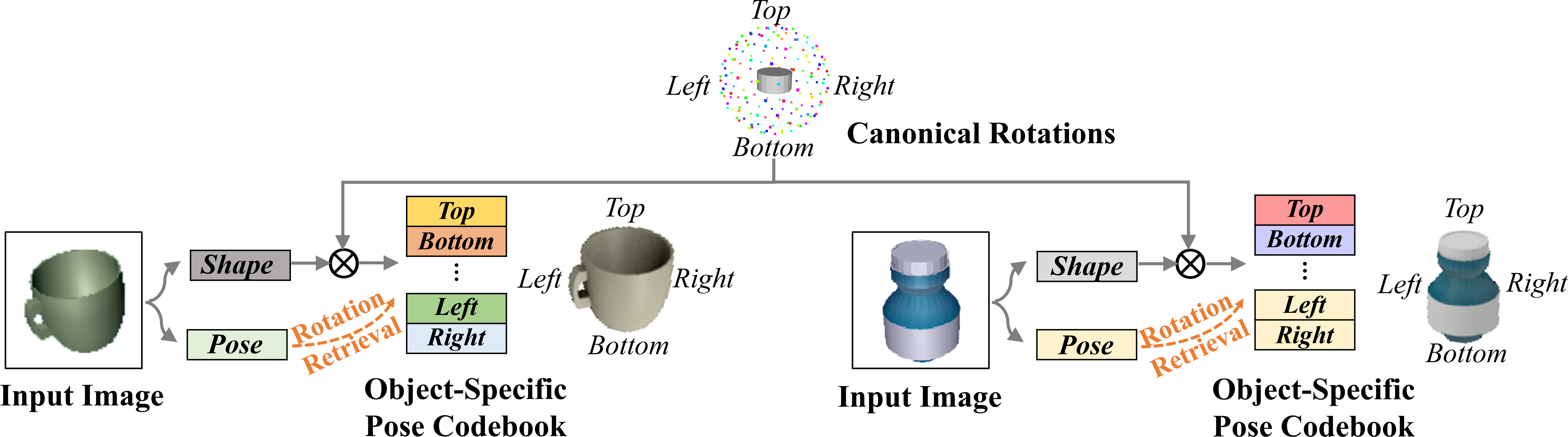}
	\caption{\textbf{Disentanglement for pose estimation.} Images of objects are mapped to latent representations for object shape and pose, respectively. Due to different object symmetries, query pose codes must refer to object-specific pose codebooks (symmetries marked by code color) for rotation retrieval, which are generated by re-entangling canonical rotations with object shapes.}
	\label{fig:motivate}
\end{figure}

Recent works improve the capability of a single network for processing multiple objects through different ways.
For example, a series of works \cite{NOCS2019,Tian2020ECCV,chen2021sgpa} perform category-level pose estimation, by learning to map input pixels (and point clouds) to corresponding points of a canonically aligned object, and computing pose registration based on the correspondences.
However, these works assume that the space of canonically aligned objects for a given category is sufficiently regular to learn with neural networks, which does not hold for different objects across categories. 
Moreover, the point-wise correspondences are ambiguous under object symmetries, which may hinder the performance of these methods.
On the other hand, Multipath-AAE \cite{multiAAE2020} builds on the auto-encoding framework \cite{sundermeyer2018implicit,sundermeyer2019augmented} to learn pose embeddings for different objects, by using a specific decoder for each object.
Therefore Multipath-AAE is not restricted by the categorical shape alignment regularity, yet the network complexity becomes prohibitive as the number of training objects gets large.
In addition, the single latent representation encoding mixed information of diverse objects under different poses may not be sufficiently accurate for pose estimation.

\begin{figure*}[t]
	\centering
	\includegraphics[width=\linewidth]{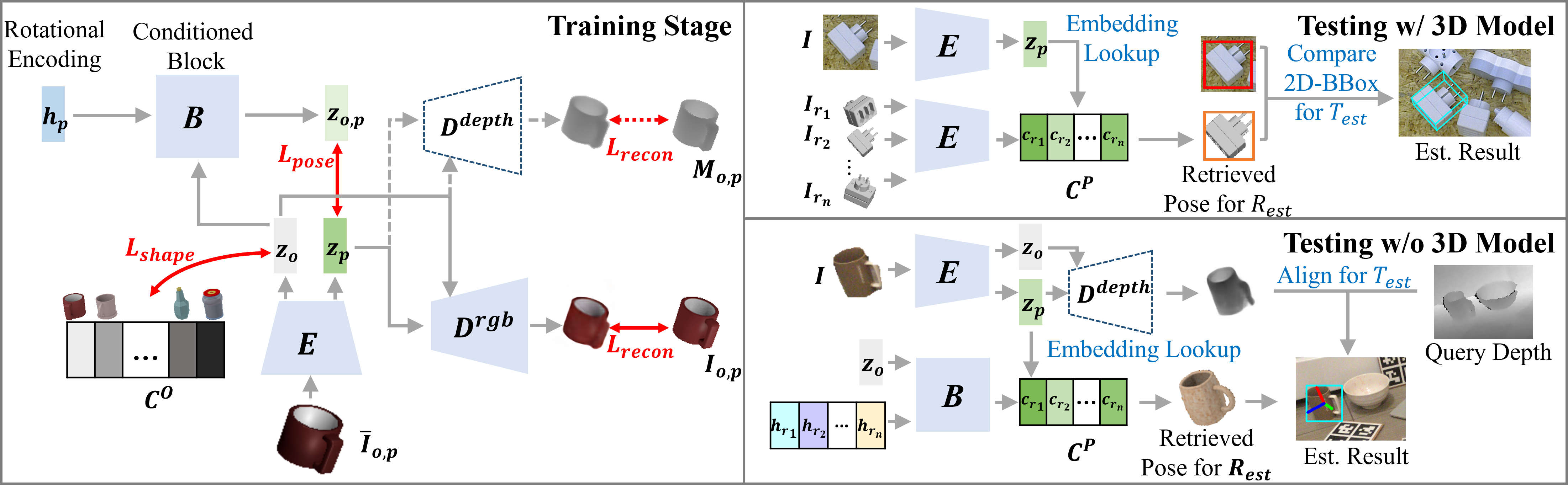}
	\caption{\textbf{Network structures} in the training (left) and testing stage (right) for different settings. If testing objects have available 3D models (or not), we train an RGB decoder only (or plus a depth decoder) (left). During test stage, object rotation is \emph{purely RGB-based estimation} by retrieving from the codebook $\mathcal{C}^P$, which is constructed by encoding the given object views (top right), or by shape code conditioned generation (bottom right). Translation is computed by pinhole camera (top right) or by depth comparison (bottom right).}
	\label{fig:framework}
\end{figure*}

We present DISP6D $-$ an approach to train a single network that processes more objects simultaneously (Fig.~\ref{fig:framework}).
As we build on the auto-encoding framework \cite{sundermeyer2019augmented}, objects do not need category labels and the symmetry ambiguity is automatically handled.
Meanwhile, we extend \cite{sundermeyer2019augmented} by \textit{disentangling} object shape and pose in the latent representation; therefore we avoid per-object decoders and reduce the network training complexity significantly. The disentanglement allows the latent pose code of an arbitrary object to be compared with a pose codebook indexed by canonical rotations for retrieval of the object rotation (see Figs.~\ref{fig:motivate},~\ref{fig:framework}), where the learned latent poses are more accurate for RGB-based pose estimation than codes mixing shape and pose information.

Learning such a disentangled representation faces a critical challenge: the different symmetries of objects do not admit one pose codebook applicable to all objects.
To understand this difficulty, consider that the cup in Fig.~\ref{fig:motivate} has distinctive codes for representing the left and right views, but the rotational symmetry of the bottle demands an identical code for the two views.
This exemplifies the frequent infeasibility of disentangling an input image into \textit{independent} latent factors by a neural network \cite{bouchacourt2021addressing}, the factors being shape and pose in our case.

To solve this dependent disentanglement problem, we model the shape-pose dependency by introducing a module that \textit{re-entangles} the shape and rotation and generates an object-conditioned pose codebook respecting the object symmetry, against which the query latent pose code is compared for pose retrieval. 
In addition, to facilitate generalization to novel objects, we take advantage of the decoupled latent shape space and apply contrastive metric learning, which encourages objects with similar geometry to have similar shape codes.  By training the system with diverse shapes, novel objects can be robustly processed by referring to similar training objects with proximate latent shape codes.

We evaluate our approach by training on synthetic data only and testing on real data. 
Our approach allows for evaluations of two different settings proposed by previous works, i.e., the textureless CAD objects without category labels proposed by \cite{multiAAE2020} and the daily objects with specified categories by \cite{NOCS2019}, on which we compare favorably than state-of-the-art methods that similarly work with RGB images for rotation estimation.
In addition, we extend to a more challenging setting of daily objects without leveraging the category information by mixing the objects from \cite{NOCS2019}, on which our approach preserves competitive performance.
These results demonstrate the improved scalability of our method.
Finally, extensive ablation studies confirm the effectiveness of disentangled shape and pose learning and other design choices.

\section{Related Works}
\label{sec:related}
\noindent\textbf{6D Pose Estimation} There is a massive literature on instance-level pose estimation from RGB(D) images (see \cite{lepetit2020survey} for a survey).
These works can be roughly classified into three streams, i.e., by direct pose regression \cite{xiang2017posecnn,kehl2017ssd,billings2018silhonet}, by registering 2D and 3D points \cite{brachmann2016uncertainty,rad2017bb8,tekin2018real,peng2019pvnet,park2019pix2pose,hodan2020epos,li2019cdpn}, and by template retrieval \cite{hinterstoisser2012model,sundermeyer2018implicit,sundermeyer2019augmented,zhang2019detect,Wen2020}. 
For instance-level pose estimation, learning-based methods train a specialized network for each testing object. 

Wang \textit{et al.}~\cite{NOCS2019} propose a shared 3D shape space (NOCS) for all instances from the same category, where the objects are pre-aligned and normalized into a common coordinate system.
Variations among the instances in the NOCS space are expected to be smooth and predictable, to make the NOCS mapping learnable when trained on large scale categorical datasets like ShapeNet~\cite{chang2015shapenet}.
For pose estimation, the pixels of a detected object are mapped to 3D points in the NOCS space, which are registered with the input depth image to find the 6D rigid transformation along with scaling.
Grabner \textit{et al.}~\cite{grabner2019location} use a similar canonical object coordinate representation for category level 3D model retrieval.

Subsequent works improve the categorical pipeline by modeling the shape differences inside a category adaptively, with many of them fusing depth with RGB input for more accurate translation and scale estimation \cite{Chen2020ECCV,Tian2020ECCV,Chen_2020_CVPR,chen2021sgpa,chen2021fs,lin2021dualposenet}.
Specifically, within the RGB-input domain, Chen \textit{et al.}~\cite{Chen2020ECCV} propose an analysis-by-synthesis approach to minimize the difference between the input image and a 2D object view synthesized by neural rendering, by gradient descent on both shape and pose variables. All these category-level approaches train different network branches for each category to learn and utilize the intra-category shape consistency.

In comparison, our scalable approach can accommodate categories of different symmetries with a common network path that learns the inter- and intra-categorical features adaptively (Fig.~\ref{fig:motivate}). Similarly, StarMap~\cite{zhou2018starmap} and PoseContrast~\cite{xiao2021posecontrast} work on the cross-category setting for estimating only the 3D rotation; however, they do not address object symmetries. LatentFusion~\cite{park2019latentfusion} does not assume categorical objects either, but requires multiple view images for neural reconstruction before pose estimation.

Multipath-AAE~\cite{multiAAE2020} works under a different assumption: the novel test objects share little shape consistency with training objects but have 3D models available, which is practical for industrial manufacturing settings. 
Multipath-AAE extends the augmented auto-encoder approach \cite{sundermeyer2019augmented} by sharing an encoder to learn the latent pose embedding and assigning to each object a separate decoder, which bypasses the large shape differences across objects and enables auto-encoding. 
The shared encoder therefore learns pose-aware features that generalize to different objects.
This setting is followed by Pitteri \textit{et al.}~\cite{pitteri2020} who use learned local surface embedding for pose estimation, and Nguyen \textit{et al.}~\cite{Nguyen_2022_CVPR} who improve robustness by modeling occlusion.
Compared with \cite{multiAAE2020}, our disentanglement of shape and pose allows the auto-encoding without multi-path decoders for different objects, thus making the framework more scalable.
However, the disentanglement into independent factors is challenging to learn and we propose re-entanglement to generate shape conditioned pose codebook for feasible learning.

\noindent\textbf{Disentangled Representation Learning}
Disentangled representations are a key objective for interpretable and generalizable learning \cite{RepLearning2013,FairnessDisentangle2019}.
Previous works encourage disentangled representation learning by unsupervised learning \cite{betaVAE17,infoGAN2016}.
Recently, focus has been given to the conditions under which learned representations can be disentangled \cite{CommonAssumptions2019,higgins2018towards,VAEICA2020}, with the finding that quite frequently the direct mapping to disentangled independent factors is unattainable for neural networks \cite{bouchacourt2021addressing}.
Our discussion on scalable 6D pose estimation exemplifies the situation: the disentanglement of object shape and pose as independent factors is prevented by different object symmetries.
We provide a solution to the disentanglement problem by re-entangling the independent factors so that a neural network mapping can be learned.

\section{Method}
\label{sec:method}

As shown in Fig.~\ref{fig:framework}, our overall framework is an auto-encoder that learns to encode an RGB image of the observed object to its latent shape code and object-dependent pose code separately, where the latent pose code is compared with a codebook of implicit rotation representations for fast pose estimation.
Therefore, our approach obtains the object rotation purely from RGB input; depth input and reconstruction are optionally used only to remove translation/scale ambiguity when the object size is unknown (Sec.~\ref{sec:inference}).

\subsection{Disentangled Shape and Pose Learning} 
\label{sec:reconstruction}

Given the input RGB image $\vb{I}_{o,p} \in \mathbb{R}^{3{\times}H{\times}W}$ for an object $o$ under pose $p\in\mathrm{SE}(3)$, the encoder $E$ maps $\vb{I}_{o,p}$ to a low-dimensional latent code $E(\vb{I}_{o,p}) = (\vb*{z}_o,\vb*{z}_p) \in \mathbb{R}^{2d}$ with $d \ll H{\times}W$, where $\vb*{z}_o,\vb*{z}_p \in \mathbb{R}^d$ encode the implicit shape and pose representations, respectively.

The decoder $D^{rgb}$ tries to recover the input image from latent codes.
Since we expect $\vb*{z}_o$ and $\vb*{z}_p$ to encode the overall object appearance and the view-specific appearance respectively, we borrow ideas from generative models \cite{dumoulin2018feature,karras2019style,karras2020styleganv2} and use the AdaIN modulation \cite{huang2017arbitrary} in the decoder to condition the per-view reconstruction on the object code; the detailed decoder structure can be found in the supplementary document. 
Moreover, we have tested by switching the roles of $\vb*{z}_o$ and $\vb*{z}_p$ for the decoder and found degraded performance (see supplemental).

Since we use only synthetic data for training, to narrow the domain gap between synthetic and real data, we follow  \cite{multiAAE2020,sundermeyer2018implicit} and adopt data augmentations that randomly change the color and scaling of an input image $\vb{I}$ to obtain the augmented image $\bar{\vb{I}}$, and aim to recover the canonical image $\vb{I}$ by auto-encoding.
The loss function of the auto-encoding task therefore is
\begin{equation}\label{eq:loss_recon}
L_{recon}=\sum_{o,p}||\vb{I}_{o,p}-D^{rgb}(E(\bar{\vb{I}}_{o,p}))||^2.
\end{equation}

Note that our design accommodates different objects by sharing the same pair of encoder-decoder $E$ and $D^{rgb}$, and hence is different from \cite{multiAAE2020} that assigns to each object an individual decoder and previous instance-level approaches that train a specialized network for each object.

\subsection{Contrastive Metric Learning for Object Shapes}\label{sec:shape}

The key to the generalization of pose estimation to a novel object is to exploit its similarity with the training objects, so that its generated pose codebook (Sec.~\ref{sec:re-entanglement}) can capture its symmetry by referring to that of similar training objects.
To learn such similarity relationships, we build a metric space for the shape codes of training objects by contrastive metric learning \cite{wu2018unsupervised,oord2018representation,he2019momentum,chen2020simple}.

Denote the training object set as $\mathcal{O}=\{o_i\}_{i\in[N_O]}$, where $N_O$ is the number of training objects. 
Similar to \cite{wu2018unsupervised}, to learn the contrastive metric among shape codes, we establish a shape embedding $\mathcal{C}^O\in\mathbb{R}^{N_O\times d}$ containing codes $\{\vb*{c}_{i}\in\mathbb{R}^d\}_{i\in[N_O]}$, each corresponding to a training object. 
We then define the proximity of $\vb*{c}_{i}$ to $\vb*{z}_o$ in the form of probability distribution as
\begin{equation}
\Pr(\vb*{c}_i|\vb*{z}_o)=\frac{\exp(\vu*{c}_i \cdot \vu*{z}_o/\tau)}{\sum_{j=1}^{N} \exp(\vu*{c}_j\cdot \vu*{z}_o/\tau)}
\label{eq:oprior1}
\end{equation}
where $\tau = 0.07$ is a temperature parameter controlling the sharpness of the distribution, and $\vu*{a}=\frac{\vb*{a}}{\norm{\vb*{a}}}$ denotes normalized unit-length vectors. 

The target distribution given $o$ is simply a one-hot vector $\vb*{w}^o \in \{0,1\}^{N_O}$, with $\vb*{w}^o_{i} = 1$ if $o=o_i$ and the rest entries being zero. 
The contrastive metric loss for learning the shape space is then defined as
\begin{equation}\label{eq:loss_shape}
L_{shape}=-\sum_{o,p}{\sum_{i=1}^{N_O}{ \vb*{w}^o_i \log{\Pr(\vb*{c}_i|\vb*{z}_o)}}}.
\end{equation}
To minimize the above loss, while $\vb*{z}_o$ is updated by the SGD solver during each training step, we update the shape embedding $\mathcal{C}^O$ by the exponential moving average (EMA) with decay rate $d_s$, thus making $\vb*{c}_o$ a smoothed history of $\vb*{z_o}$.
Details of the EMA update can be found in the supplementary document.

\subsection{Re-entanglement of Shape and Pose}
\label{sec:re-entanglement}

The pose code $\vb*{z}_p$ is compared with a codebook of sampled canonical orientations to retrieve the object rotation (Figs.~\ref{fig:motivate},~\ref{fig:framework}).
As noted in Fig.~\ref{fig:motivate}, different object symmetries demand object-specific pose codebooks.
To generate such a conditioned pose codebook, we propose a distributed representation of rotations and a transformation that entangles rotations with shape code in a generalizable way.

\noindent\textbf{Rotational Position Encoding}
We need to distinguish between different rotations in a canonical pose representation.
Inspired by the positional encoding in sequence models \cite{transformer}, we have adopted the 4D hyper spherical harmonics (HSH) rotation encoding.
The HSH is a set of orthogonal basis functions on the 4D hypersphere that mimic the sine/cosine wave functions for positional encoding in sequence models: it is a distributed vector representation that can extend to high dimensions ($d=128$ in our case), has a multi-spectrum structure that encodes both high frequency and low frequency variations of rotations, and has periodic structures with fixed linear transformations for relative rotations \cite{zhao2017spherical,4dhsh2015}.
Denoting the HSH function as $Z_{nl}^{m}(\beta,\theta,\phi)$, with $\beta\in[0,2\pi]$, $\theta\in[0,\pi]$, $\phi\in[0,2\pi]$ as the in-plane rotation, zenith and azimuth angles respectively and $l,m,n$ as polynomial degrees, we obtain the 128-dim vector encoding $\vb*{h}_p$ by ranging over $n\in[0,\cdots,6]$ with $0\leq l \leq n$, $0\leq m \leq l$.
Details of the construction can be found in the supplemental document.

\noindent\textbf{Conditioned Pose Code Generation} We design a conditional block $B$ to entangle the object code $\vb*{z}_o$ with the rotational position encoding $\vb*{h}_p$ of rotation $p$ and output a pose code $\vb*{z}_{o,p} = B(\vb*{z}_o, \vb*{h}_p)$ comparable with $\vb*{z}_p$ (Fig.~\ref{fig:framework}).

Entanglement is a recurring topic in machine learning, with implementation techniques like parameter generation \cite{platanios2018contextual,tian2020conditional,dumoulin2018feature} that boil down to a tensor product structure \cite{Tenenbaum2000,martyn2020entanglement}.
Therefore, we introduce a 3rd-order learnable tensor $\vb{W}\in{\mathbb{R}^{d\times d\times d}}$ and apply the following two-step transformation $B$ to obtain the entangled pose code:
\begin{align}
\vb*{z}'_{o,p} = \vb{W}\left({FC}(sg(\vb*{z}_o)), {FC}(\vb*{h}_p)\right), \quad 
\vb*{z}_{o,p} = \texttt{FFN}(\vb*{z}'_{o,p}), 
\label{eq:bt}
\end{align} 
where ${FC}(sg(\vb*{z}_o)), {FC}(\vb*{h}_p) \in \mathbb{R}^{d}$ are the pre-processing of $\vb*{z}_o$ and $\vb*{h}_p$,  $sg(\cdot)$ is to stop gradient back-propagation as the shape code $\vb*{z}_o$ is a pre-condition not to be updated by pose learning (see Sec.~\ref{sec:ablation}, Tab.~\ref{table:real275_pose_ablation} for an ablation), and $\vb{W}(\cdot,\cdot)$ denotes the tensor contraction along its first two orders.
A feed-forward residual block \texttt{FFN} is followed to generate the final pose code $\vb*{z}_{o,p}$.

To synchronize the pose representation computed via the conditional block with that learned by the encoder, we minimize the cosine distance between $\vb*{z}_{o,p}$ and $\vb*{z}_p$ during training:
\begin{equation}\label{eq:loss_pose}
L_{pose} = -\sum_{o, p}{\vu*{z}_{o,p}\cdot\vu*{z}_p}.
\end{equation}

In summary, our total training loss combines the reconstruction loss (Eq.~(\ref{eq:loss_recon})), the contrastive loss for shape space (Eq.~(\ref{eq:loss_shape})) and the synchronization loss between pose representations from $B$ and $E$ (Eq.~(\ref{eq:loss_pose})), with weights $\lambda_1,\lambda_2$:
\begin{equation*}
    L=L_{recon}+\lambda_1 L_{shape}+\lambda_2 L_{pose}.
\end{equation*}

\section{Inference under Different Settings}
\label{sec:inference}

In the test stage, we estimate rotation purely from RGB input, which takes three steps (Fig.~\ref{fig:framework}, right): Given the query image crop $\vb{I}$ bounding the object of interest, we first obtain its latent shape and pose codes as $(\vb*{z}_o, \vb*{z}_p) = E(\vb{I})$, then build a pose embedding $\mathcal{C}^P {\in} \mathbb{R}^{N_P{\times} d}$ with each row $\vb*{c}_q{\in}\mathbb{R}^d$ corresponding to the rotation $q$ from a set of $N_P$ canonical rotations $\mathcal{R} {\subset} SO(3)$, and finally retrieve the estimated pose as $q^* = \argmax_{q\in \mathcal{R}}{\vu*{z}_p\cdot\vu*{c}_q}$. Translation (and scale) is estimated subsequently, which may use depth data to remove scale ambiguity.

Previous works on scalable pose estimation towards novel objects have assumed two different application scenarios as discussed below, on which our framework can be flexibly adapted and achieve state-of-the-art performances. We also present an extended setting to better explore the scalability of our approach.

\noindent\textbf{Setting I: Novel Objects in a Given Category}
A series of works \cite{NOCS2019,Chen2020ECCV,Tian2020ECCV} assume that the novel testing objects are from a specific category but have no 3D models available.
Therefore, for pose retrieval we compute $\mathcal{C}^P=\{B(\vb*{z}_o, \vb*{h}_q)\}_{q\in\mathcal{R}}$ from the sampled canonical rotations $\mathcal{R}$ and the shape code $\vb*{z}_o$.

As the testing objects have no specific sizes in this setting, to remove the 2D-3D scale ambiguity and estimate translation and scale properly, we require the input depth map and compare it with a decoded canonical depth map.
The estimation of translation and scale involves a simple outlier point removal process and mean depth comparison for translation estimation and bounding box comparison for scale estimation; for details please refer to the supplemental document.
As shown in Fig.~\ref{fig:framework}, the depth decoder $D^{depth}$ is simply an additional branch parallel to the RGB decoder, supervised to reconstruct a canonical depth map $\vb{M}_{o,p}\in\mathbb{R}^{1\times H\times W}$ for the rotated object at a fixed distance away from the camera. 
The reconstruction loss in Eq.~(\ref{eq:loss_recon}) is updated to be:

\begin{equation}
    L_{recon}=\sum_{o,p}{||\vb{I}_{o,p} - D^{rgb}(E(\bar{\vb{I}}_{o,p}))||^2  + ||\vb{M}_{o,p} - D^{depth}(E(\bar{\vb{I}}_{o,p}))||^2 }
    \label{eq:recon_wdepth}
\end{equation}
Comparison in Sec.~\ref{sec:comparison_unknown} shows our improved rotation accuracy and robustness to object symmetries.

\noindent\textbf{Setting II: Novel Objects with 3D Models}
Multipath-AAE~\cite{multiAAE2020} works with a set of CAD objects with drastic geometric differences and no specific category consistency.
However, the 3D models of novel testing objects are accessible, as is common in applications like industrial manufacturing \cite{multiAAE2020,pitteri2020}. 

In this setting, we follow previous auto-encoding frameworks \cite{multiAAE2020,sundermeyer2019augmented} to construct an offline pose codebook with the CAD model. Specifically, we first render images $\vb{I}_q$ of the given object under the reference orientations $q$ and then obtain $\mathcal{C}^P = \{\vb*{z}_q\}_{q\in\mathcal{R}}$, with $\vb*{z}_q$ the pose code part of $E(\vb{I}_q)$. 
Given the physical size and camera intrinsics, translation is obtained purely from RGB input with the pinhole camera model. The decoder $D^{rgb}$ is not used during the test stage.
As shown in Sec.~\ref{sec:comparison_known}, our disentangled auto-encoder learns highly discriminative pose encoding that performs even better than per-object trained auto-encoders, and generalizes well to novel objects with largely different shapes.

\noindent\textbf{Setting III (Extension): Novel Objects across Categories without 3D Models}
We further challenge our method on an extension of setting I by combining objects of all categories in \cite{NOCS2019} into one set.
Without referring to predefined category labels in training and testing, the task has never been addressed before in previous works\cite{NOCS2019,Chen2020ECCV,Tian2020ECCV}.
As shown in Sec.~\ref{sec:ours_all}, our disentangled auto-encoder enables a straightforward extension to this cross-category setting with marginal performance degrading compared to setting I, which demonstrates the scalability of our approach.

\section{Experiments} 
\label{sec:experiment}

\subsection{Setup}\label{sec:training_setup}

We resize the input images to $H{\times}W = 128{\times}128$, use a latent code dimension $d = 128$, and set $d_s=0.9995$ for the EMA decay, $\lambda_1 = 0.004,\lambda_2 = 0.002$ for balancing the loss terms.
We use the Adam optimizer~\cite{kingma2014adam} with default parameters and a learning rate of 0.0002, and train 50k iterations for settings I, II, and 150k iterations for setting III, with a batch size of 64 to convergence. Detailed network structure and training data preparation are in the supplementary document.

\begin{figure}[tb]
\begin{minipage}[t]{.45\textwidth}
\centering
\resizebox{1.\textwidth}{!}{
\begin{tabular}{ccccc}
    \hline 
    &  \tabincell{c}{NOCS\\ \protect\cite{NOCS2019}} & \tabincell{c}{Chen et al. \\ \protect\cite{Chen2020ECCV}} & \tabincell{c}{Ours-\\per} &\tabincell{c}{Ours-\\all} \\
    \hline
\tabincell{c}{Only synthetic \\training data} & $\times$  & $\checkmark$ & $\checkmark$ & $\checkmark$\\ 
\hline
\tabincell{c}{Only RGB for \\ rotation est.} & $\times$ & $\checkmark$ & $\checkmark$ & $\checkmark$\\
\hline
\tabincell{c}{Extension to \\cross-category} & $\times$ & $\times$& $\times$ & $\checkmark$ \\
    \hline
\end{tabular}}
\end{minipage}
\hfill
\begin{minipage}[t]{.54\textwidth}
\centering
    \includegraphics[width=1.\textwidth]{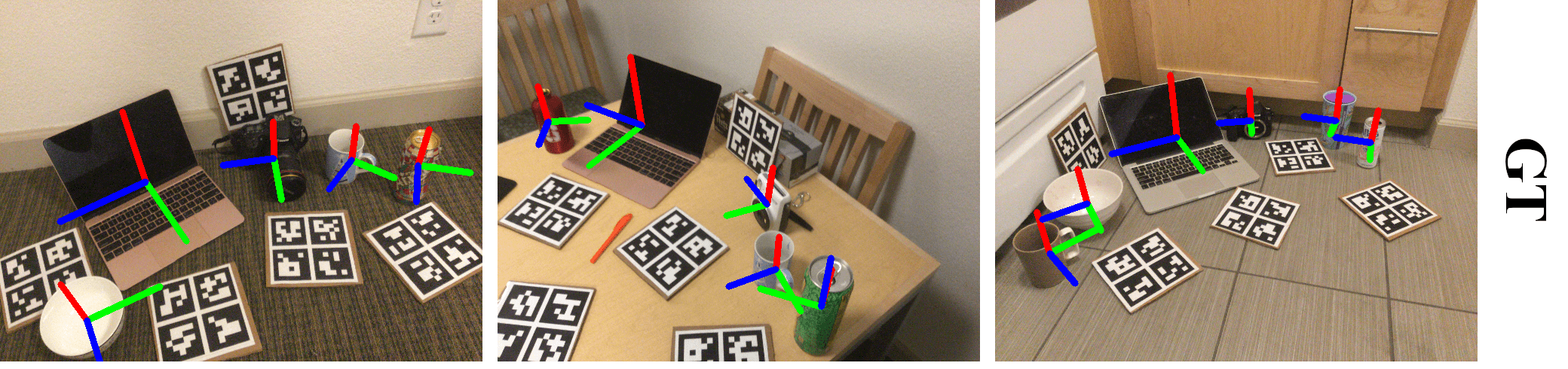}
    \vfill
    \includegraphics[width=1.\textwidth]{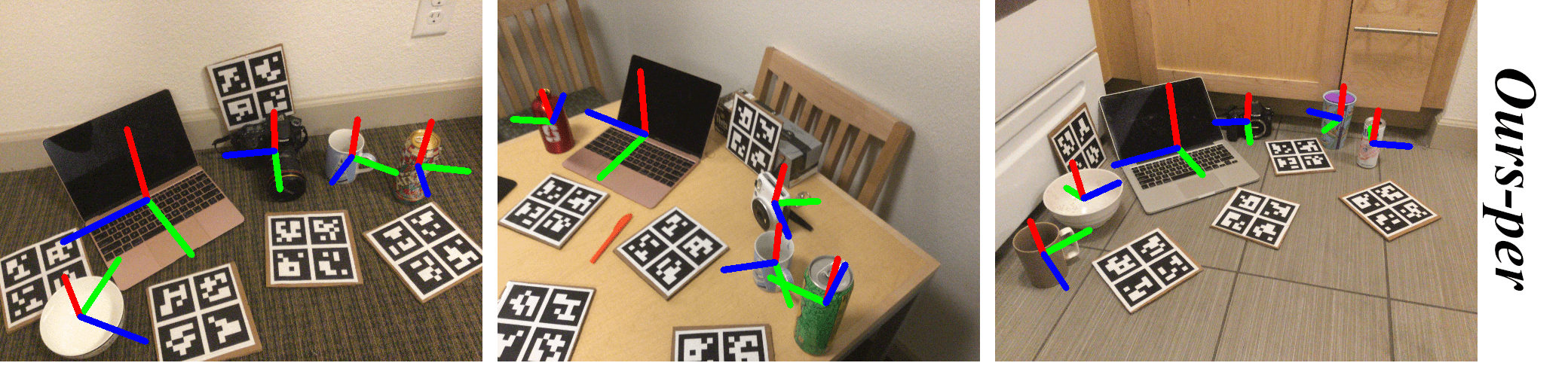}
\end{minipage}
\caption{Scope of compared methods on settings I and III (left), and qualitative cases of \textit{Ours-per} (right). \emph{All methods use query depth for translation estimation.}}
\label{fig:real275_setting1}
\end{figure}

\subsection{Setting I: Novel Objects in a Given Category}
\label{sec:comparison_unknown}

\noindent\textbf{Dataset and Metrics } The benchmark of \cite{NOCS2019} has two parts, i.e., CAMERA containing synthetic data and REAL275 containing real data, that span 6 categories of objects (\textit{bottle, bowl, camera, can, laptop, mug}) situated in daily indoor scenes.
Furthermore, the objects in a category have diverse scales, and due to the inherent 2D-3D scale ambiguity, the estimation of translation plus scaling is only possible when additional cues like depth are given.

We use the synthetic CAMERA dataset with 1085 objects for training and evaluate on the real test set of REAL275, and follow \cite{Chen2020ECCV} to report the average precision (AP) at different thresholds of rotation and translation errors.
Note that while \cite{Chen2020ECCV} uses input depth for improved translation estimation, it assumes a fixed scale and thus does not address scale estimation. Nevertheless, for completeness we report our scale estimation result by measuring 3D IoU precision in the supplemental document.

\noindent\textbf{Baselines } 
The most relevant baseline is \cite{Chen2020ECCV}, as both methods train on synthetic data only and test on real data, and estimate rotation based on RGB input only and use depth only for translation estimation.
Another baseline is the earlier \cite{NOCS2019}, which however trains on both real and synthetic data and relies on input depth for rotation estimation.
All three methods use the same 2D detection backbone Mask-RCNN adopted from \cite{NOCS2019}.
We summarize the differences in scopes of three methods in Fig.~\ref{fig:real275_setting1}(left) where our method in this setting is denoted \textit{Ours-per}, and defer an empirical discussion of more category-level methods taking RGB-D input for rotation estimation \cite{Chen_2020_CVPR,Tian2020ECCV,lin2021dualposenet,chen2021sgpa} to the supplemental.

\noindent\textbf{Pose Codebook} 5K reference rotations are obtained by K-means clustering on the CAMERA training set rotations. Generating a pose codebook from 5K HSH codes takes 0.04s on a GTX 1080 GPU and can be batched for more objects.

\begin{figure*}[tb]
 \centering
    \includegraphics[width=.49\linewidth]{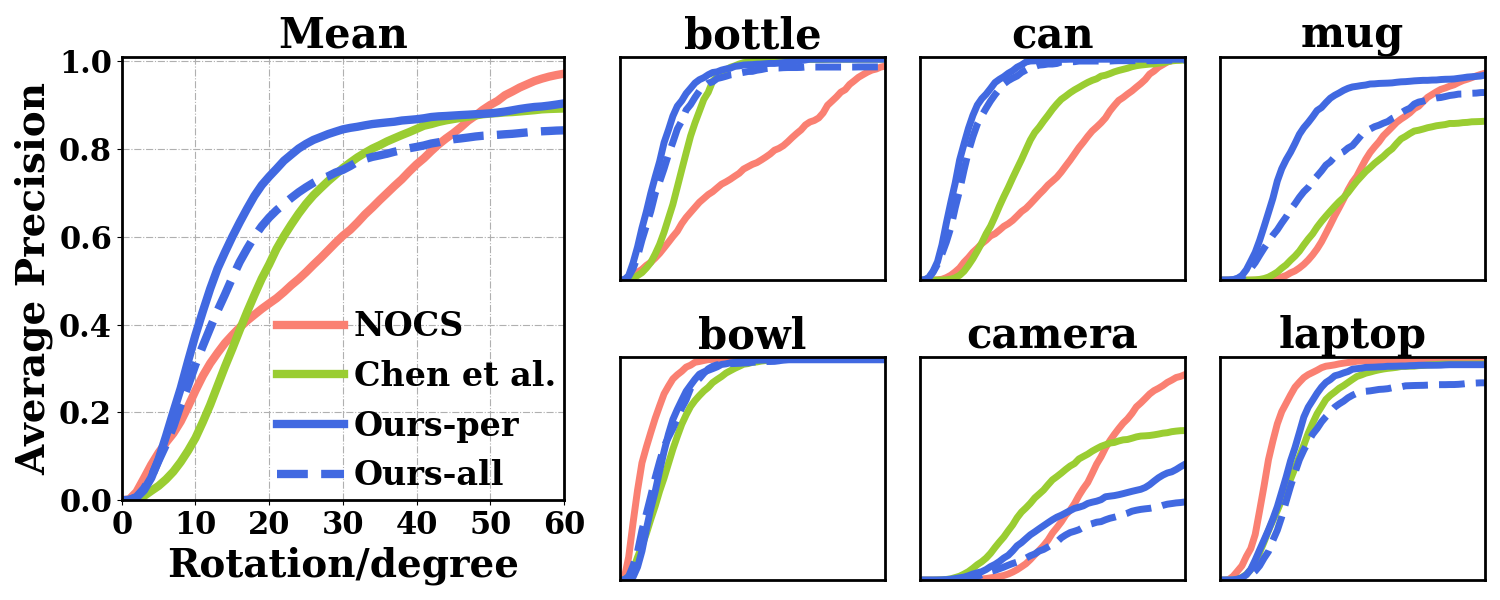}
 \hfill
    \includegraphics[width=.49\linewidth]{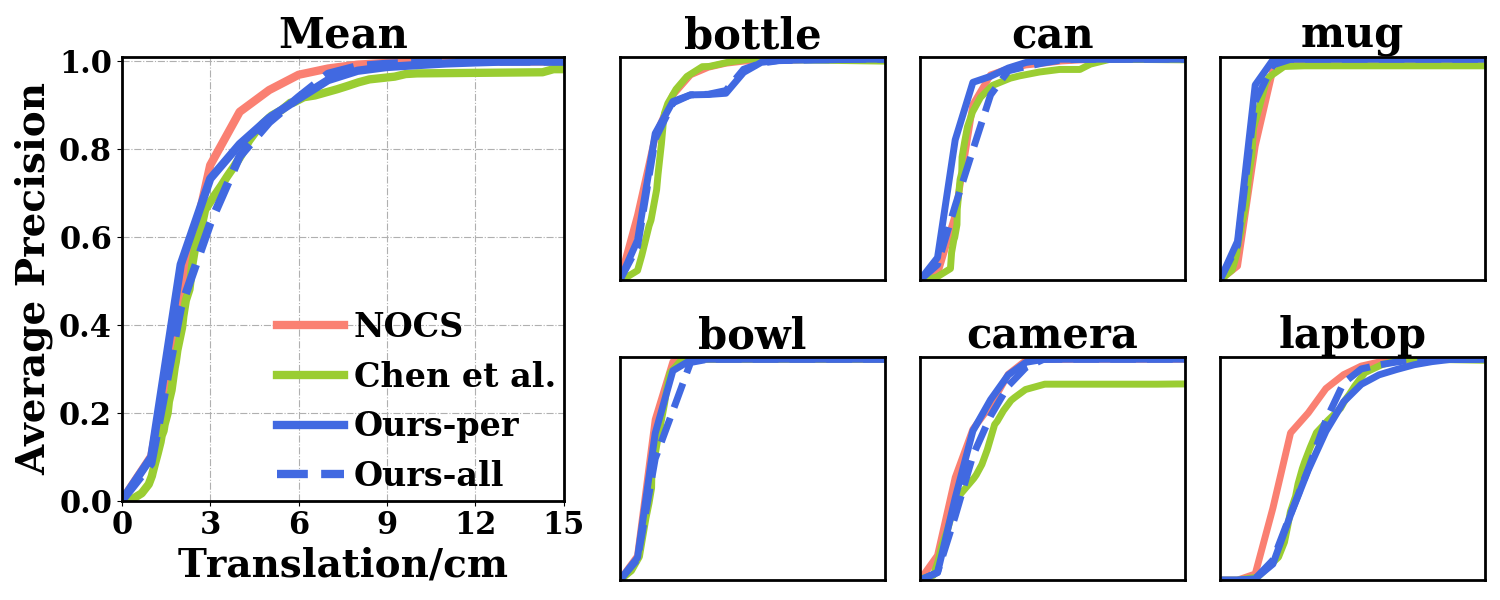}
 \caption{\textbf{Comparison on REAL275} of average precision (ranging from $0$ to $1$) at different rotation error (left, ranging from $0$ to $60^{\circ}$) or translation error (right, ranging from $0$ to $15cm$) thresholds. We report \textit{Ours-per} of setting I per-category level and \textit{Ours-all} of setting III combining all 6 categories.}
 \label{fig:real275}
\end{figure*}

\noindent\textbf{Results } As shown in Fig.~\ref{fig:real275}, compared with Chen \textit{et al.}\cite{Chen2020ECCV}, our rotation estimation has increased AP with a significant margin when the error threshold is below $40^{\circ}$; meanwhile, both methods have comparable performances on translation estimation. 
Compared with NOCS\cite{NOCS2019}, our margin is even more significant throughout the range of $10^{\circ}$ to $45^{\circ}$ for rotation estimation.
Qualitative results are visualized in Fig.~\ref{fig:real275_setting1}. 
Among the different categories, we perform better in the classes of bottle, can and mug, which have strong partial symmetries and our method handles robustly.
However, the camera category poses difficulty to our method; the main reason is that subtle textures are needed to distinguish vastly different poses, e.g., the front and back of a camera are quite similar for flat lens, but there are few objects out of the totally 74 objects in training set to cover such texture diversities. 
In comparison, both \cite{Chen2020ECCV} and \cite{NOCS2019} use optimization to search for rotation and are more resilient to severe train/test disparities. 
For scale estimation, our 3D IoU accuracy is comparable to \cite{NOCS2019} (see supplemental).

\subsection{Setting III (Extension): Novel Objects across Categories without 3D Models}\label{sec:ours_all}
We further challenge our method on the extended setting that combines all 6 categories of the NOCS benchmark into one set, without referring to category labels in training and testing; the trained network is denoted \textit{Ours-all}. As we learn a metric shape space without the need for category labels (Sec.~\ref{sec:shape}), we expect our method to extend to this cross-category setting without much difficulty.

As shown in Fig. \ref{fig:real275}, for rotation estimation, \textit{Ours-all} achieves improved results than Chen \textit{et al.}\cite{Chen2020ECCV} for error thresholds $<28^\circ$, and NOCS \cite{NOCS2019} for error thresholds in $10^{\circ}\sim40^{\circ}$, even though \cite{Chen2020ECCV,NOCS2019} train per-category network modules to exploit the intra-category consistency. 
Meanwhile, the lower performance compared with \textit{Ours-per} can be attributed to the confusion of shape-conditioned pose learning introduced by the increased cross-category shape variances, as for example under certain views a mug with an occluded handle looks quite similar to a can or bowl, but they are forced to generate pose codes with different symmetries. Qualitative cases are given in the supplemental.

Although none of the previous works \cite{Chen2020ECCV,NOCS2019} are designed to address this setting, for a better understanding of the challenge, we adapt and retrain NOCS \cite{NOCS2019} by using a single head for all categories (\textit{i.e., NOCS-all}); without per-category correspondence consistency, we find that \textit{NOCS-all} performs poorly especially for rotation estimation. 
We also retrain PoseContrast~\cite{xiao2021posecontrast} under our setting, which is the state-of-the-art for cross-category rotation estimation. Results show that ~\cite{xiao2021posecontrast} does not handle objects with different symmetries as well as we do. Details are given in the supplemental.

\subsection{Setting II: Novel Objects with 3D Models}
\label{sec:comparison_known}

\noindent\textbf{Dataset and Metrics }  
Following \cite{multiAAE2020}, we evaluate on T-LESS \cite{hodan2017tless} which contains 30 textureless industrial parts with very different shapes and symmetries (see the supplementary for a visualization).
Accuracy is measured by the recall rate of visible surface discrepancy metric $e_{VSD} < 0.3$ \cite{hodan2018bop} at distance tolerance $20mm$, among test instances with visible portion ${>}10\%$.

\noindent\textbf{Baselines } We compare with  Multipath-AAE\cite{multiAAE2020}, Pitteri \textit{et al.}~\cite{pitteri2020}, and Nguyen \textit{et al.}~\cite{Nguyen_2022_CVPR}. All these methods share the same setting by training jointly on only the first 18 objects and testing on all 30 objects, using CAD models from TLESS.

\noindent\textbf{Pose Codebook } We follow \cite{sundermeyer2018implicit,sundermeyer2019augmented,multiAAE2020} to build for each test object an offline pose codebook with $92232$ reference rotations, that is formed by combining 36 in-plane rotations and 2562 equidistant spherical views sampled via \cite{hinterstoisser2008simultaneous}.

\noindent\textbf{Results }
We first report in Tab.~\ref{tab:tless_combined}(a) the accuracy for all test instances with 2D GT bounding boxes. 
We outperform Multipath-AAE \cite{multiAAE2020} by 4\% on average for the novel objects (\textit{i.e.,} Obj 19-30) and 5\% for the trained objects (\textit{i.e.,} Obj 1-18), although Multipath-AAE \cite{multiAAE2020} assigns separate decoders for the 18 training objects and optionally uses the GT mask to eliminate background noise for better performances. We also outperform the concurrent work by Nyugen \textit{et al.}~\cite{Nguyen_2022_CVPR}.
For a more complete evaluation, we further compare with \cite{sundermeyer2018implicit,sundermeyer2019augmented} which train for each of the 30 objects a specific auto-encoder, and find our result still outperforms it by 3\% on the 18 training objects of ours. 
These results show that our disentanglement learning improves the auto-encoder framework and generalizes to objects with different shapes and symmetries (see Sec.~\ref{sec:ablation}, Fig.~\ref{fig:ablation_pose} for detailed analysis).

We then report in Tab.~\ref{tab:tless_combined}(b) the evaluation under the full 2D detection and pose estimation pipeline, by adopting Mask-RCNN \cite{he2017mask} from \cite{labbe2020cosypose} as the 2D detector and following the single object single instance protocol \cite{hodan2018bop}. Our result improves over that of the comparing methods by a significant margin of around 12\%.  Our qualitative cases are in Fig.~\ref{fig:qualitative_tless} and the per-object recall rates are given in the supplementary.

\begin{table}[tb]
\centering
\caption{\textbf{Comparison on T-LESS.} Reported are the average recall rates with $e_{VSD}<0.3$.  All methods were trained with only the first 18 objects, except AAE\cite{sundermeyer2018implicit,sundermeyer2019augmented} which trains individual networks for each of the 30 objects.}
\label{tab:tless_combined}
\begin{subtable}[t]{0.59\textwidth}
\begin{center}
\caption{w/ 2D GT bboxes, $\dag$ for using GT mask}
    \label{table:tless_GT}
\resizebox{1.0\textwidth}{!}{
\begin{tabular}{cccc}
    \hline 
    \textbf{Ave. on} & \textbf{Obj 1-18}& \textbf{Obj 19-30}&\textbf{Obj 1-30}\\
    \hline
    AAE\cite{sundermeyer2018implicit,sundermeyer2019augmented} & 62.57 & \textbf{66.63} & 64.19 \\
    \hline
    Multipath-AAE\cite{multiAAE2020} & 51.75 &52.49 &52.04\\
    Multipath-AAE\cite{multiAAE2020}$^\dag$& 60.75 &59.89 &60.41 \\
    Nguyen \textit{et al.}\cite{Nguyen_2022_CVPR}  & 59.62  & 57.75 & 58.87\\
    Ours & \textbf{66.14} & \textbf{64.42} & \textbf{65.45} \\
    \hline
\end{tabular}}
\end{center}
\end{subtable}
\hfill
\begin{subtable}[t]{0.39\textwidth}
\begin{center}
\caption{w/ MaskRCNN \cite{he2017mask} detection.}\label{table:tless_2D}
\resizebox{.95\textwidth}{!}{
\begin{tabular}{cc}

\hline
  \textbf{Ave. on }& \textbf{Obj 1-30} \\
 \hline
 Multipath-AAE\cite{multiAAE2020} & 23.51\\
 Pitteri \textit{et al.}~\cite{pitteri2020} & 23.27\\
 Ours & \textbf{35.36}\\
\hline
\end{tabular}}
\end{center}
\end{subtable}
\end{table}

\begin{figure}[tb]
 \centering
 \includegraphics[width=1.\textwidth]{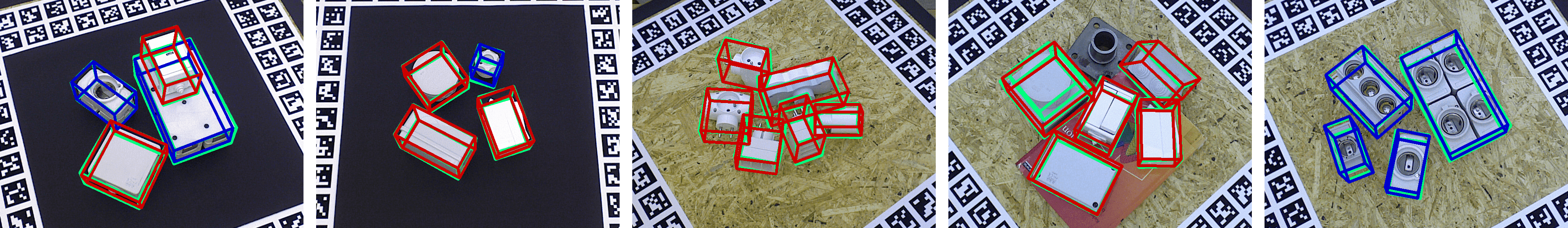}
 \caption{\textbf{Qualitative results on T-LESS} of setting II. We denote our estimations in blue (trained objects) and red (unseen objects), and GT poses in green.}
 \label{fig:qualitative_tless}
\end{figure}

\noindent\textbf{Instance-Level Estimation}
Although we focus on scalable pose estimation for novel test objects, it is possible to apply our framework to an instance-level task where all test objects are given for training. 
We provide such a limit case study in the supplementary, and compare with more instance-level pose estimation methods on the BOP leader board \cite{sundermeyer2018implicit,sundermeyer2019augmented,park2019pix2pose,li2019cdpn,labbe2020cosypose,hodan2020epos}. Our approach provides fast yet accurate pose estimations that can be further improved by refinement.

\subsection{Ablation Study}\label{sec:ablation}
\noindent\textbf{Shape Conditioned Pose Code Generation} We first discuss the necessity to generate shape-dependent pose codes. To this end, we separate shape codes from pose codebook generation by replacing the 3rd-order tensor $\vb{W}$ in Eq.~(\ref{eq:bt}) with a multi-layer perceptron $\texttt{MLP}$ that takes only the HSH encoding as input, \textit{i.e.} $\texttt{MLP}({FC}(\vb*{h}_p))$. The $\texttt{MLP}$ has four layers of width $[1024, 1024, 1024, 128]$ and thus more trainable weights than $\vb{W}$.
The average precision on setting III reported in Tab.~\ref{table:real275_pose_ablation} (2nd, 6th rows) shows that the performance significantly drops when the shape code is separated from pose code generation, indicating the difficulty of learning independent latent representations of shape and pose.

To further visualize the effectiveness of pose code generation, given an object $o$, we inspect two sets of latent pose representations: $\mathcal{C}^P_E=\{\vb*{z}_p\}_{p\in\mathcal{R}}$ generated by the encoder $E$ and $\mathcal{C}^P_B=\{\vb*{z}_{o,p}\}_{p\in\mathcal{R}}$ by the conditioned block $B$. $\mathcal{R}$ has 8020 rotations from a combination of 20 in-plane rotations and 401 quasi-equidistant views sampled via \cite{gonzalez2010measurement}.
Ideally, the two sets of latent codes should coincide with each other, so that they can be compared for effective rotation estimation.

\begin{figure}[tb]
 \centering
 \begin{subfigure}[b]{0.49\textwidth}
    \centering
    \includegraphics[width=1.\linewidth]{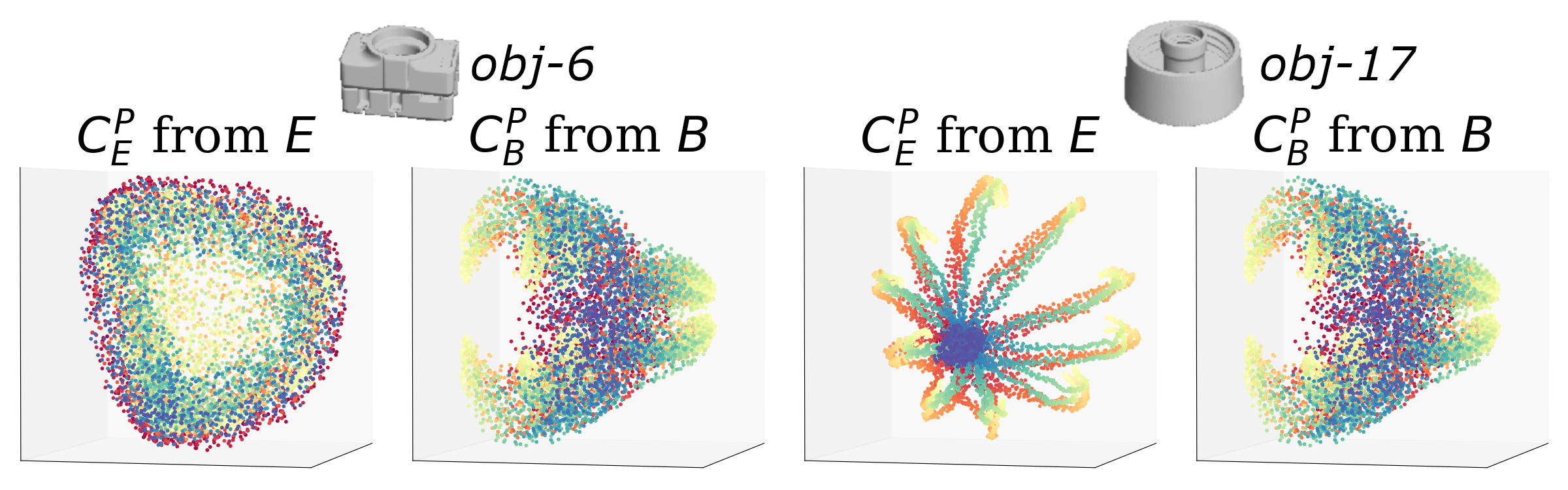}
    \caption{w/o shape condition}
 \end{subfigure}
 \hfill
 \begin{subfigure}[b]{0.49\textwidth}
    \centering
    \includegraphics[width=1.\linewidth]{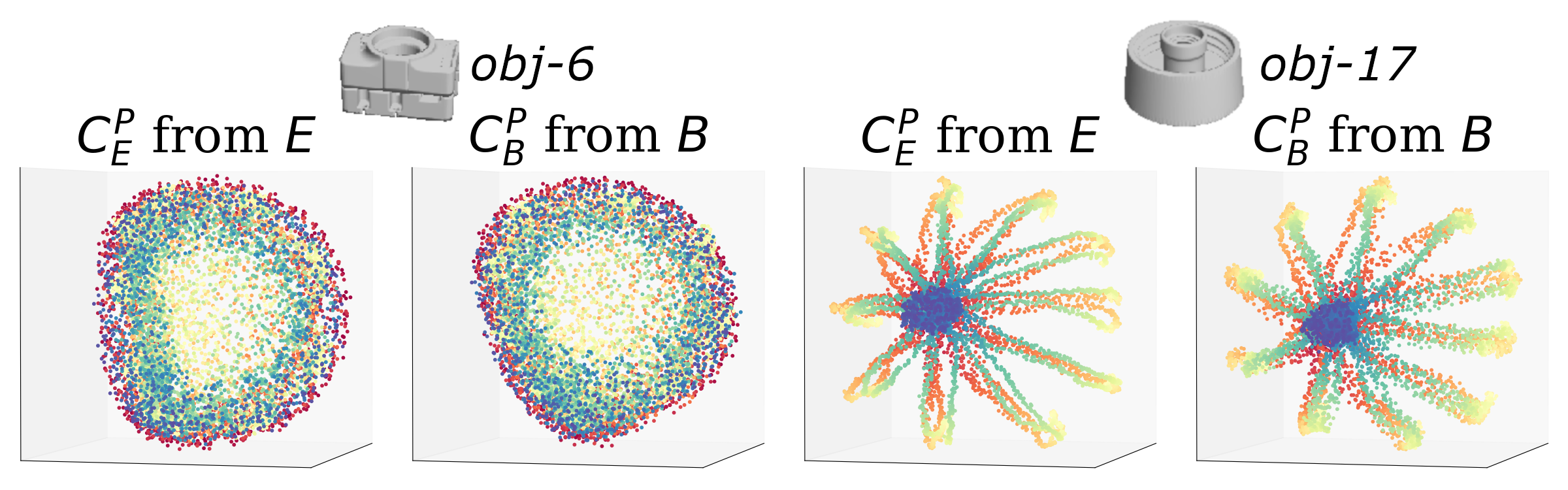}
    \caption{w/ shape condition}
 \end{subfigure}
 \caption{\textbf{Top three PCA projections of pose codes $\mathcal{C}^P_E$ and $\mathcal{C}^P_B$} from encoder $E$ and condition block $B$ for two T-LESS objects. Point colors (\textcolor{blue}{blue}$\rightarrow$\textcolor{green}{green}$\rightarrow$\textcolor{red}{red}) encode rotations as viewpoints change from north pole to south pole. The shape conditioned pose codes well capture the symmetries and synchronize with encoder outputs (b), but unconditioned pose codes fail (a).}
 \label{fig:ablation_pose}
\end{figure}

We show in Fig.~\ref{fig:ablation_pose} for two T-LESS training objects: the box-like Obj-6 and the cylinder-like Obj-17, where with our entanglement of shape and pose information, $\mathcal{C}^P_B$ well synchronize with $\mathcal{C}^P_E$ for objects with different degrees of symmetry, though for Obj-6 a global rotation of the PCA projections between $\mathcal{C}^P_B$ and $\mathcal{C}^P_E$ exists due to the nearly isotropic distribution of latent codes. On the contrary, when the shape code is isolated from generating the pose codebook, it becomes difficult for $\mathcal{C}^P_B$ to follow the pattern of $\mathcal{C}^P_E$ for different objects. Such contrast demonstrates the necessity of our entanglement.
We further discuss in the supplementary for objects with texture solving the rotational ambiguity, where our pose codes can well capture the textural difference.

\begin{table}[t]
\centering
\caption{\textbf{Ablation tests} on the design of shape conditioned pose code generation and contrastive learning for object shape. Reported are mAP at different rotation error thresholds (in degrees) for mixed categories of REAL275 (setting III).}
\label{table:real275_pose_ablation}
\label{table:real275_shape_ablation}
\begin{center}
\small
\resizebox{0.8\textwidth}{!}{
\begin{tabular}{p{4cm}<{\centering}ccccccc}
    \hline 
    Design of $B$& w/ $L_{shape}$ & $\textit{AP}_5$ & $\textit{AP}_{10}$ & $\textit{AP}_{15}$ & $\textit{AP}_{20}$ & $\textit{AP}_{30}$ & $\textit{AP}_{60}$\\
    \hline
    $\texttt{MLP}({FC}(\vb*{h}_p))$ & $\checkmark$& 4.7 & 15.7 & 28.9 & 36.9 & 47.0 & 72.4 \\
    $\texttt{MLP}({FC}(sg(\vb*{z}_o)), {FC}(\vb*{h}_p))$  & $\checkmark$ & 7.5 & 27.3 & 47.8 & 61.8 & 74.9 & 84.3\\
    $\vb{W}\left({FC}(\vb*{z}_o), {FC}(\vb*{h}_p)\right)$ & $\checkmark$ & 6.6 & 26.6 & 47.8 & 62.0 & \textbf{76.6} & \textbf{87.5}\\
    $\vb{W}\left({FC}(sg(\vb*{z}_o)), {FC}(\vb*{h}_p)\right)$ & $\times$ & 2.8 & 15.2 & 33.6 & 48.9 & 67.6 & 81.3 \\
    $\vb{W}\left({FC}(sg(\vb*{z}_o)), {FC}(\vb*{h}_p)\right)$& $\checkmark$  &  \textbf{9.1} & \textbf{30.9} & \textbf{50.7} & \textbf{64.4} & 75.3 & 84.3\\
    \hline
\end{tabular}}
\end{center}
\end{table}

We then move on to validate the design of combining pose and shape. An intuitive idea is to simply concatenate the shape and pose rotational encoding and process by an MLP, \textit{i.e.} $\texttt{MLP}({FC}(sg(\vb*{z}_o)), {FC}(\vb*{h}_p))$, with \texttt{MLP} having four layers of width $[1024, 1024, 1024, 128]$. The comparison in Tab.~\ref{table:real275_pose_ablation} (3rd, 6th rows) shows that the 3rd-order tensor outperforms $\texttt{MLP}$, thus verifying our design choice.

Finally, we validate the necessity to treat $\vb*{z}_o$ as a pre-condition for pose code generation, by allowing gradients to be backpropagated through the conditioned pose code generation module to $\vb*{z}_o$ instead. 
Tab.~\ref{table:real275_pose_ablation}, 4th and 6th rows, show that pre-conditioning by stop gradient $sg(\vb*{z}_o)$ performs better for rotation error thresholds ${\leq}20^\circ$, demonstrating its recognition of subtle pose differences.

\begin{figure}[tb]
    \includegraphics[width=.49\linewidth]{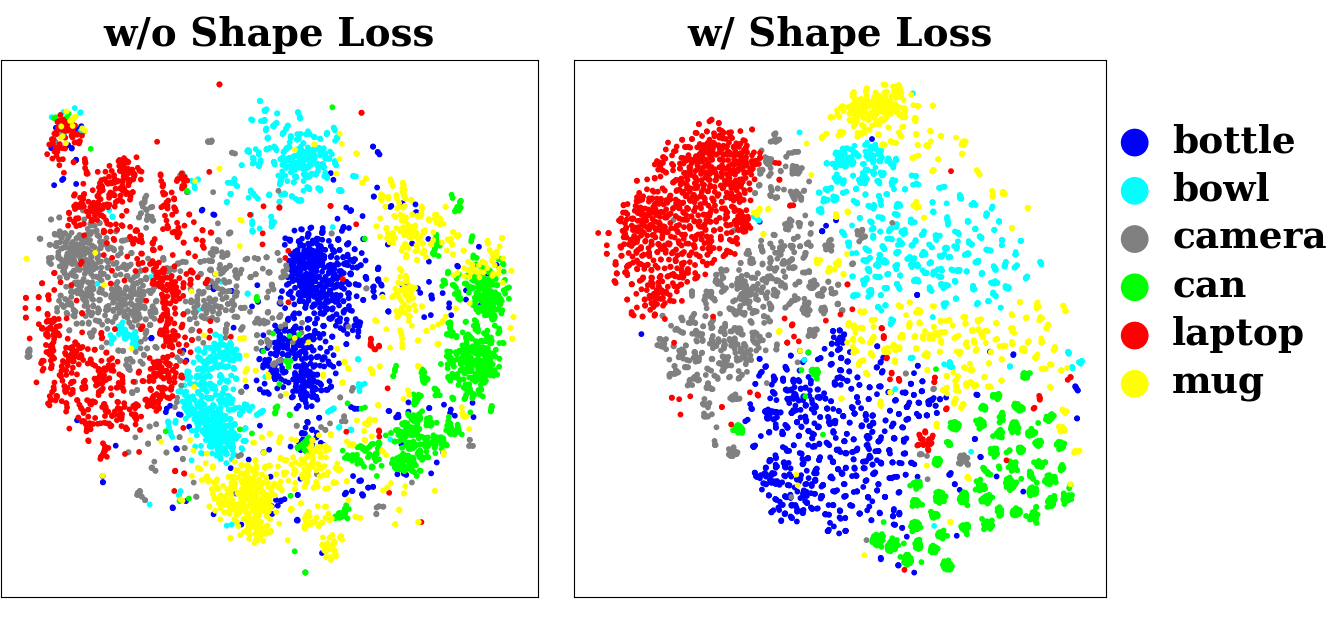}
    \hfill
    \includegraphics[width=.49\linewidth]{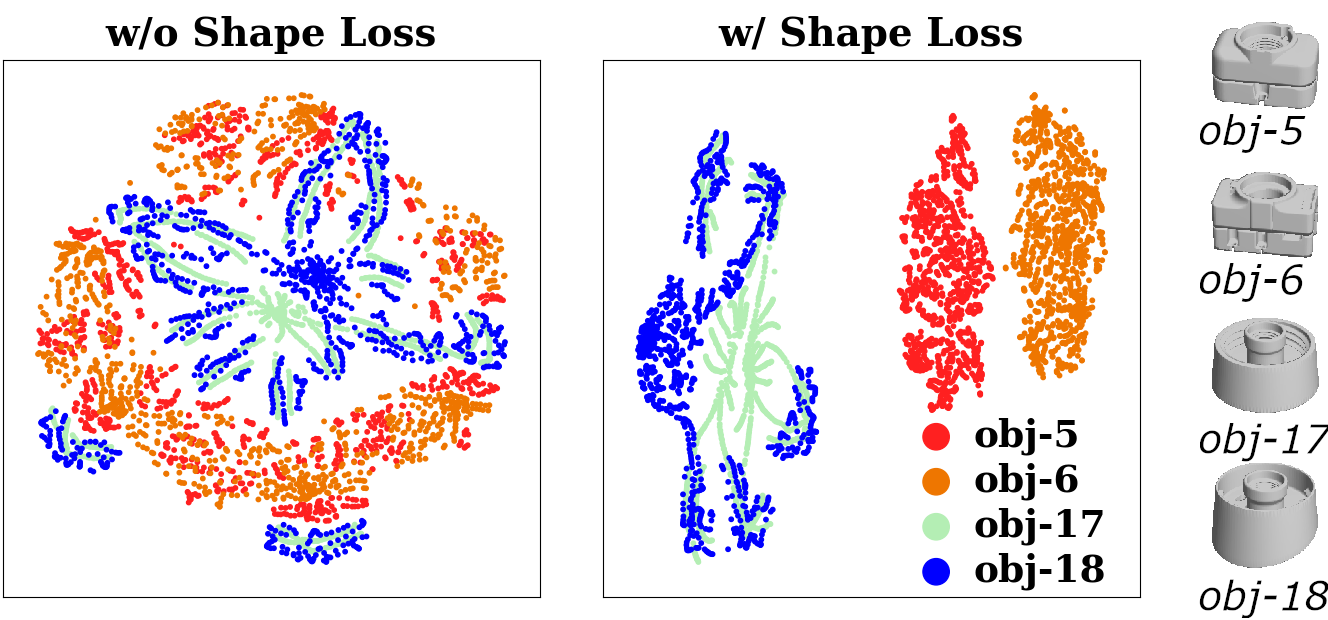}
 \caption{\textbf{t-SNE embedding of shape codes $\vb*{z}_o$} for training images of six CAMERA categories (left) and four T-LESS objects (right). With contrastive metric learning the shape spaces show better regularity w.r.t. shape similarities.}
 \label{fig:ablation_obj}
\end{figure}

\noindent\textbf{Contrastive Metric Learning for Object Shapes }
The mAP in Tab.~\ref{table:real275_shape_ablation} (5th, 6th rows) demonstrates our gain from the contrastive metric learning of the shape space, where with the shape loss $L_{shape}$ the generalization to unseen objects is significantly improved.
We also visualize the shape codes $\vb*{z}_o$ with t-SNE in Fig.~\ref{fig:ablation_obj}, for training samples from the CAMERA objects and 4 T-LESS objects.
With shape space metric learning, we observe much better intra-category clustering and inter-category separation on CAMERA, though the network is unaware of category labels in this setting (setting III). For the T-LESS objects, the introduction of $L_{shape}$ not only well separates the box-like objects (Obj-5,6) from the cylinder-like objects (Obj-17,18), but also recognizes the detailed geometric differences between Obj-5 and Obj-6;
in comparison, the shape codes for different objects are mixed together without shape space metric learning.

\section{Conclusion}
\label{sec:conclusion}

We have presented a simple yet scalable approach for 6D pose estimation that generalizes to novel objects unseen during training. 
Building on an auto-encoding framework that handles object symmetry robustly, we achieve scalability by disentangling the latent code into shape and pose representations, where the shape representation forms a metric space by contrastive learning to accommodate novel objects, and the pose code is compared with canonical rotations for pose estimation. As disentanglement into independent shape and pose spaces is fundamentally difficult due to different object symmetries, we re-entangle shape code with pose codebook generation to avoid the issue. We obtain state-of-the-art results on two established settings when training with synthetic data only, and extend to a cross-category setting to further demonstrate scalability.

\noindent\textbf{Limitation and Future Work} 
We mainly focus on learning for rotation estimation from a single RGB image, while the translation estimation can be further improved by fully exploiting the input depth with neural networks, as discussed in \cite{Tian2020ECCV,lin2021dualposenet}. Extending to multiview input for improved robustness under severe occlusion and inaccurate 2D detection is also a promising direction.

\noindent\textbf{Acknowledgement} This work was partially supported by the Innovation and Technology Commission of the HKSAR Government under the InnoHK initiative.


\clearpage
\setcounter{page}{1}
\setcounter{table}{0}
\setcounter{figure}{0}

\appendix
\section{Overview}
The supplementary document is divided into several sections, to provide more details of our design and discussion of experiments mentioned in the main text:

Sec.~\ref{sec:supp_network} presents the detailed network structures, including our AdaIN modulation with ablation on the decoder design by switching the roles of shape and pose codes, as well as the 4D HSH formula for rotational position encoding.

Sec.~\ref{sec:supp_training} includes additional details about the training procedure, including the EMA update of $\mathcal{C}^O$ and training data synthesis. 

Sec.~\ref{sec:supp_tra} illustrates how we conduct translation and scale estimation for novel objects with unknown physical size (Settings I, III).

Sec.~\ref{sec:supp_oursper} supplements the discussion for \textit{Ours-per} (Setting I), with more qualitative results, comparison with NOCS under the 3D IoU metric, and comparison with RGBD fusion networks.

Sec.~\ref{sec:supp_oursall} supplements the discussion for \textit{Ours-all} (Setting III), with qualitative results, comparison with \textit{NOCS-all} and PoseContrast, and the visualization of pose codes for a wine bottle with texture resolving its axial symmetric ambiguity.

Sec.~\ref{sec:supp_tless} supplements discussion for setting II on T-LESS,  with the visualization of 30 T-LESS models, more qualitative cases and our per-object recall rate for ``detection+pose estimation" pipeline (Tab. ~\ref{table:tless_2D} of main text). 

Sec.~\ref{sec:supp_instlevel} reports our evaluation regarding the instance-level pose estimation.

\begin{figure*}[tb]
    \centering
    \includegraphics[width=0.95\linewidth]{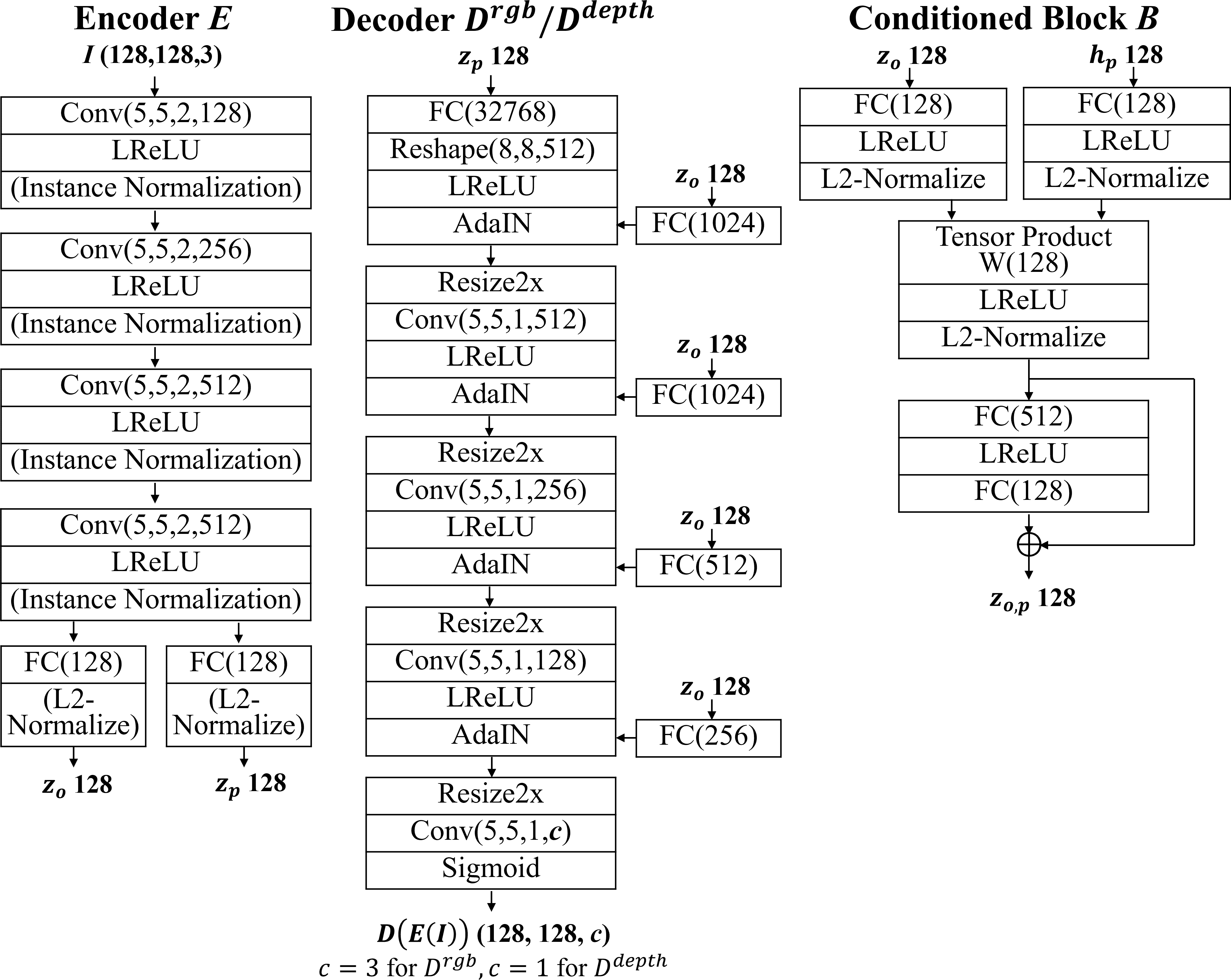}
    \caption{The detailed network structures. The layer parameters are in the formats: Conv(\textit{filter height, filter width, stride, filter number}), FC(\textit{output dimension}) and W(\textit{output dimension}). For encoder, we use instance normalization (IN) and L2 normalization only when training on the NOCS CAMERA dataset.}
    \label{fig:detail_network}
\end{figure*}

\section{Detailed Network Structures}\label{sec:supp_network}
We provide the detailed network architectures of encoder $E$, decoder $D^{rgb}$ and $D^{depth}$, and conditioned block $B$ in Fig.~\ref{fig:detail_network}. Furthermore, we explain details of the AdaIN modulation and 4D HSH formula for rotational position encoding in the subsections below.

\subsection{AdaIN Modulation (Sec.~\ref{sec:reconstruction} of Main Text)}\label{sec:supp_adain}
We use the AdaIN modulation\cite{huang2017arbitrary} to condition the per-view reconstruction on object code.
Specifically, we transform the shape code by $(\vb*{g}_i^s, \vb*{g}_i^b) = FC_i(\vb*{z}_o) \in \mathbb{R}^{2C_i}$ and modulate the intermediate feature map $\vb{F}_i \in \mathbb{R}^{C_i\times H_i \times W_i}$ decoded from the pose code by
\begin{equation*}
\widetilde{\vb{F}}_i = \vb*{g}^s_i \odot \frac{\vb{F}_i-\mu(\vb{F}_i)}{\sigma(\vb{F}_i)} + \vb*{g}^b_i,
\end{equation*}
where $FC_i$ is a fully connected layer, $\odot$ is the element-wise product, and $\mu(\cdot)$ and $\sigma(\cdot)$ compute the mean and standard deviation vectors across the spatial dimensions, respectively.

We note that an alternative is to swap the roles of shape $\vb*{z}_o$ and pose $\vb*{z}_p$, and decode from $\vb*{z}_o$ with $\vb*{z}_p$ as the condition in AdaIN modulation. We compare these two configurations by following setting III (Sec.~\ref{sec:ours_all} of main text) to train on the CAMERA by combining all 6 categories into one set, and report the average precision on REAL275 in Tab.~\ref{table:dec}.

Tab.~\ref{table:dec} verifies our design of using the $\vb*{z}_o$-conditioned decoder: the swapping of $\vb*{z}_o$ and $\vb*{z}_p$ for decoder notably degrades performance across different thresholds.

\begin{figure}[tb]
 \centering
 \begin{subfigure}[b]{1.\textwidth}
    \centering
    \includegraphics[width=.7\linewidth]{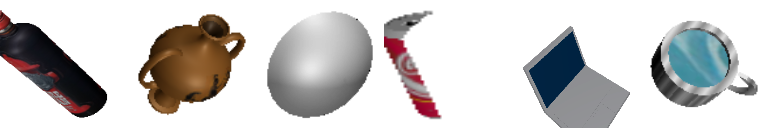}
    \caption*{Query image crops from the CAMERA training set}
 \end{subfigure}
 \vfill
 \begin{subfigure}[b]{1.\textwidth}
    \centering
    \includegraphics[width=.7\linewidth]{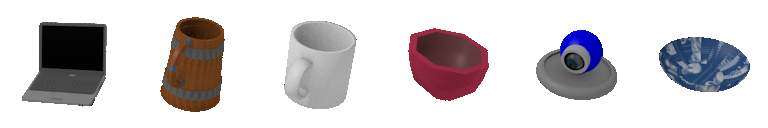}
    \caption*{Retrieved 1NN object, w.r.t using a $\vb*{z}_p$-conditioned decoder}
 \end{subfigure}
 \vfill 
 \begin{subfigure}[b]{1.\textwidth}
    \centering
    \includegraphics[width=.7\linewidth]{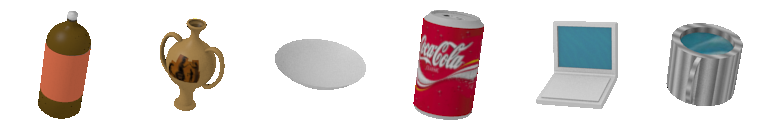}
    \caption*{Retrieved 1NN object, w.r.t using a $\vb*{z}_o$-conditioned decoder}
 \end{subfigure}
 \caption{Qualitative results on shape retrieval for image crops from the CAMERA training set, with regard to different decoder designs. For each column, the top row shows the training image, the middle row the retrieved 1-nearest neighbor object by using the $\vb*{z}_p$-conditioned decoder, and the bottom row the nearest neighbor object by using the $\vb*{z}_o$-conditioned decoder.}
 \label{fig:ablation_dec}
\end{figure}

Qualitatively we observe that during training with the $\vb*{z}_p$-modulated decoder, the shape contrastive loss is hard to minimize and as a result, objects cannot find their corresponding representations as nearest neighbors in the latent shape space (see Fig.~\ref{fig:ablation_dec} for examples of training images not retrieving proper objects based on cosine distance  Eq.~\ref{eq:oprior1} of main text), which also lead to poor scaling to novel objects at the test stage.
This performance difference from the $\vb*{z}_o$-modulated decoder can be attributed to the fact that AdaIN modulation changes the overall structure of an image by the spatially uniform affine transformation, which better matches the semantics of shape representation that controls drastic shape variations, rather than the semantics of pose representation that controls the gradual and local variation of viewpoints. 
By using $\vb*{z}_p$ for overall structural AdaIN modulation, the $\vb*{z}_p$-conditioned decoder effectively forces the shape code to encode both different objects and their subtle view changes simultaneously, which are conflicting aims and lead to many difficulties for learning both shape space and its conditioned pose space.

\begin{table}[tb]
\caption{Ablation study on the decoder design. Reported are mAP at different thresholds of rotation error (in degrees) for mixed categories of REAL275 (setting III)}
\centering
\begin{center}
\small
\resizebox{0.8\textwidth}{!}{
\begin{tabular}{p{4cm}<{\centering}cccccc}
    \hline 
    & $\textit{AP}_5$ & $\textit{AP}_{10}$ & $\textit{AP}_{15}$ & $\textit{AP}_{20}$ & $\textit{AP}_{30}$ & $\textit{AP}_{60}$\\
    \hline
    $\vb*{z}_p$-conditioned decoder & 2.6 & 14.7 & 29.8 & 43.5 & 59.9 & 79.1 \\
    $\vb*{z}_o$-conditioned decoder & \textbf{9.1} & \textbf{30.9} & \textbf{50.7} & \textbf{64.4} & \textbf{75.3} & \textbf{84.3}\\
    \hline
\end{tabular}}
\end{center}
\label{table:dec}
\end{table}

\subsection{4D HSH Formula (Sec.~\ref{sec:re-entanglement} of Main Text)}\label{sec:supp_hsh}
For a rotation $p\in SO(3)$ with in-plane rotation $\beta\in[0,2\pi]$, zenith $\theta\in[0,\pi]$ and azimuth $\phi\in[0,2\pi]$, and $l,m,n$ being the polynomial degrees, the corresponding 4D HSH function is constructed as
\begin{equation*}
\begin{split}
    Z^m_{nl}(\beta,\theta,\phi) = &  2^{l+1/2}\sqrt{\frac{(n+1)\Gamma(n-l+1)}{\pi\Gamma(n+l+2)}}\Gamma(l+1)\\
    &\sin^l\frac{\beta}{2} C^{l+1}_{n-l}(\cos\frac{\beta}{2})Y^m_l(\theta,\phi)\\
\end{split}
\end{equation*}
with $C^{l+1}_{n-1}$ as the Gegenbauer polynomials, and the 3D spherical harmonics $Y^m_l(\theta,\phi)\in\mathbb{C}$ as 
\begin{equation*}
Y^m_l(\theta,\phi)=\sqrt{\frac{2l+1}{4\pi}\frac{(l-m)!}{(l+m)!}}e^{im\phi}P^m_l(\cos{\theta}),
\end{equation*}
where $P^m_l$ is the associated Legendre function. 
We then follow Alg.~\ref{alg:hsh} to compute each dimension of the rotational position encoding $\vb*{h}_p$ for pose $p$. 
We have shifted $\beta$ with a small delta of $0.05\pi$, as we observe that having $\beta=0$ among the sampled rotations would lead to $\sin^l\frac{\beta}{2}=0$ for $l\neq0$ and thus the full 128 dimensions of $\vb*{h}_p$ become sparse.

\begin{algorithm}[h]
\caption{Generation of the rotational position encoding $\vb*{h}_p$ for rotation $p(\beta,\theta,\phi)$ }
\begin{algorithmic}[0]
\STATE $\beta:=\beta+0.05\pi$ \
\STATE $x:=0$\
\FOR{each $n \in [0,...,6]$}
\FOR{each $l \in [0,...,n]$}
\FOR{each $m \in [0,...,l]$}
\WHILE{the dimension index $x\leq128$}
\STATE $\vb*{h}_p[x]:=Re(Z^m_{nl}(\beta,\theta,\phi))$
\STATE $\vb*{h}_p[x+1]:=Im(Z^m_{nl}(\beta,\theta,\phi))$
\STATE $x:=x+2$
\ENDWHILE
\ENDFOR
\ENDFOR
\ENDFOR
\end{algorithmic}\label{alg:hsh}
\end{algorithm}

\section{Training Procedure}\label{sec:supp_training}
\subsection{EMA Update of $\mathcal{C}^O$ (Sec.~\ref{sec:shape} of Main Text)}
For each $\vb*{c}_i\in\mathcal{C}^O$, we maintain for it two variables: $n_i\in \mathbb{R}^+$ and $\vb*{m}_i\in\mathbb{R}^{d},d=128$, with $n_i$ initialized as 1, and each entry of $\vb*{m}_i$ randomly initialized by the normal distribution $\mathcal{N}(0,1)$. During each SGD iteration, the variables are updated as follows:
\begin{align*}
n_i &\leftarrow d_s n_i+(1-d_s)\sum_{o}\vb*{w}^o_i\\
\vb*{m}_i &\leftarrow d_s \vb*{m}_i+(1-d_s)\sum_{o}\vb*{w}^o_i\vb*{z}_o\\
\vb*{c}_i &\leftarrow \vb*{m}_i/n_i
\end{align*}
where $o$ iterates over the training objects in a mini-batch, $\vb*{w}^o\in\{0,1\}^{N_O}$ is the target distribution used for the shape contrastive metric learning, and $d_s$ is the exponential decay rate.

\subsection{Training Data and Augmentation Strategy}\label{sec:training_data}
To prepare the training images for T-LESS objects, we sample 92232 rotations from the combination of 36 in-plane rotations and 2562 equidistant spherical views sampled via \cite{hinterstoisser2008simultaneous}. 
With these sampled rotations, we follow AAE\cite{sundermeyer2018implicit} and Multipath-AAE\cite{multiAAE2020} to rotate and center the object with a fixed distance along the camera axis ($700mm$), and render the groundtruth images under fixed lighting with a plain background. Note that our rendering uses simple lighting and rasterization, rather than the physically-based renderer (PBR) in \cite{hodan2020bop,denninger2019blenderproc}.
We then augment the corresponding encoder input image following Multipath-AAE\cite{multiAAE2020}, with random operations including: 1) changing lighting conditions, 2) applying 2D translation and 2D scaling, 3) adding random background images, and 4) tone mapping the color channels.

To train on the CAMERA dataset, we take the training images of the CAMERA dataset.
For each instance we use the image patch masked by its groundtruth 2D mask as the encoder input, and further augment the image by following Multipath-AAE to: 1) apply random 2D scaling and 2) randomly adjust brightness for the \textit{camera} category and color channels for other categories. We separately process \textit{camera} and other categories, because for \textit{camera} the color hue is critical for distinguishing poses (\textit{e.g.,} the front side with lens and the back side with display). 
To prepare the corresponding reconstruction target, we place the 3D model under the groundtruth rotation and a fixed distance along the camera axis (1 $unit$), and render the target image under fixed lighting with a plain background. 
Moreover, noticing the biased tendency of aligned training cameras to have their handles on the left side, we augment the camera objects by flipping the z-coordinate of camera meshes and putting them into the training set.

\section{Translation and Scale Estimation for Novel Objects with Unknown Physical Size (Settings I, III)}\label{sec:supp_tra}
For novel test objects whose physical sizes are unknown, the RGB-based translation estimation along the camera axis becomes ill-conditioned due to the scale ambiguity. Therefore, we refer to depth to remove the scale ambiguity and estimate the translation along the camera axis (\textit{i.e., }$T_z$), by bounding box size comparison and mean depth comparison between the query depth and the depth reconstructed from $D^{depth}$. The full 3D translation $\vb*{T}=(T_x, T_y, T_z)^T$ could then be recovered with the pinhole camera model, based on the estimated $T_z$ and the detected 2D bounding box center $(t_x, t_y)$. We illustrate the detailed process below.

\noindent\textbf{Step 1.} Transform the observed query depth map $\vb{M}$ to the point cloud $\vb{O}$ in the query camera coordinate system with the query camera intrinsic $K$:
    \begin{equation*}
        \vb{O}=\mathcal{W}(K, \vb{M})
    \end{equation*}
where $\mathcal{W}(\cdot,\cdot)$ is the inverse-projection operation to transform the depth map into the 3D point cloud under the camera space. We note that the $z$-axis is the camera axis and the $z$-coordinates of the 3D point cloud are equal to the depth map values, while the $x,y$-coordinates are recovered accordingly by referring to the pinhole camera model with the given camera intrinsic.

We further filter the query $\vb{O}$ with outlier removal, where we delete points in $\vb{O}$ if they are distant from their $k$-th (\textit{e.g., }$k=100$) nearest neighbors searched in $\vb{O}$, with the distance threshold as $5cm$.
    
\noindent\textbf{Step 2.} Transform the reconstructed depth  $\vb{M}_{r}=D^{depth}(E(\vb{I}))$ to a point cloud $\vb{O}_{r}$ in the object coordinate system, where $\vb{I}$ is the query image and $\vb{O}_{r}$ is recovered to be a representation of the visible part for the observed object placed with the observed rotation in its object coordinate system. 
    
To derive $\vb{O}_{r}$ from $\vb{M}_{r}$, we recall that $\vb{M}_{r}$ is supervised to reconstruct a canonical depth map of the object under the observed orientation, whose groundtruth signal is rendered by placing the rotated object at a fixed distance $t_{r,z}$ along the camera axis (\textit{i.e.,} $z$-axis). Therefore with the training camera intrinsic $K_{r}$ and $T_{r,z}$, $\vb{O}_{r}$ could be recovered by an inverse process of rendering:
\begin{equation*}
    \vb{O}_{r}=\mathcal{W}(K_{r}, \vb{M}_{r}) - (0,0,T_{r,z})^T
\end{equation*}
    
\noindent\textbf{Step 3.} Estimate the relative scaling factor $s$ from $\vb{O}_{r}$ to $\vb{O}$. We compute $s$ as the diagonal ratio of the bounding box along the $x,y$-axes between $\vb{O}$ and $\vb{O}_{r}$. Note that for scale estimation we do not refer to the $z$-axis, in order to reduce the influence from the noise of depth values.

\noindent\textbf{Step 4.} Estimate the translation along the $z$-axis (\textit{i.e., }camera axis) $T_z$ with the mean depth comparison  
    \begin{equation*}
    T_z=\avg(\vb{O}_z)-s\avg(\vb{O}_{r,z})\label{eq:est_tz}
    \end{equation*}
where $\avg(\vb{O}_z),\avg(\vb{O}_{r,z})$ respectively denote the average of $z$-value for $\vb{O}$ and $\vb{O}_{r}$.

\noindent\textbf{Step 5.} Recover the full translation. We assume the 2D projected bounding box of decoder output is centered, and derive the translation $T_x, T_y$ along the $xy$-plane with the pinhole camera model as
\begin{equation*}
\begin{split}
    T_x=T_z(t_x- p_x)/f_x\\
    T_y=T_z(t_y- p_y)/f_y
\end{split}\label{eq:est_txy}
\end{equation*}
where $(t_x,t_y)$ is the 2D bounding box center obtained from detection, and $(f_x,f_y), (p_x,p_x)$ are the focal and principal point of the query camera intrinsic $K$.

\noindent\textbf{Step 6.} Outlier removal. We note that the mean depth and bounding box size are sensitive to outliers. Therefore we align $s\vb{O}_{r}+\vb*{T}$ with the observed $\vb{O}$ and conduct a simple outlier removal for both $\vb{O}_{r}$ and $\vb{O}$, where from  $\vb{O}_{r}$ we remove point $\vb*{p}_r \in \vb{O}_{r}$ if $s\vb*{p}_r+\vb*{T}$ is distant from the observed depth $\vb{O}$, and from $\vb{O}$ we remove points that are distant from $s\vb{O}_{r}+\vb*{T}$. 
    
\noindent\textbf{Step 7.} Update $s,\vb*{T}$ with the filtered $\vb*{O}_{r}$ and $\vb{O}$. We once again estimate the translation $\vb*{T}$ and scale $s$ by following the procedure described from Step 3 to Step 5 and comparing the filtered $\vb{O}_{r}$ and $\vb{O}$.

\section{Supplementary for \textit{Ours-per} (Sec.~\ref{sec:comparison_unknown} of Main Text, Setting I)}\label{sec:supp_oursper}

\noindent\textbf{Qualitative Results} 
We provide in Fig.~\ref{fig:supp_real275} more qualitative cases of our pose estimation result for \textit{Ours-per} on all 6 scenes of the REAL275 testing set.

\begin{figure*}[p]
 \centering
 \begin{subfigure}[b]{0.95\textwidth}
    \centering
    \includegraphics[width=1.\linewidth]{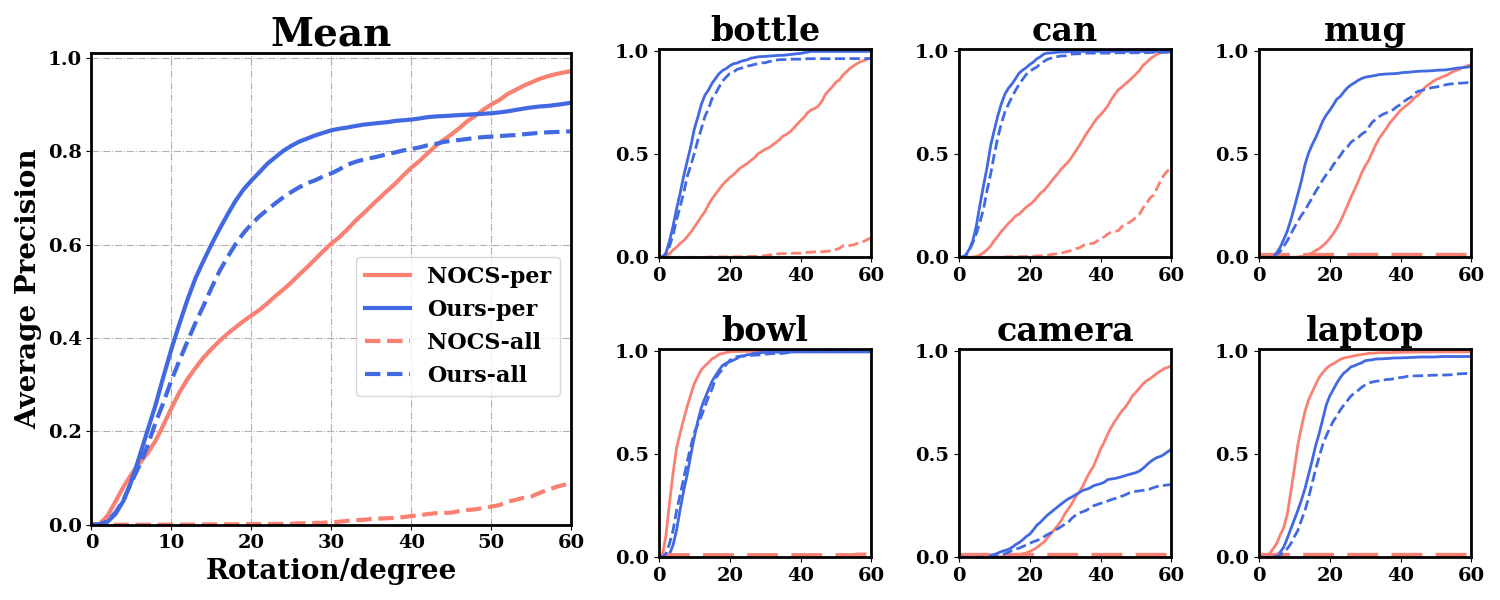}
    \caption{\textbf{Rotation}: AP at different rotation error thresholds}
    \label{fig:real275_rot_nocsall}
 \end{subfigure}
 \vfill
 \begin{subfigure}[b]{0.95\textwidth}
    \centering
    \includegraphics[width=1.\linewidth]{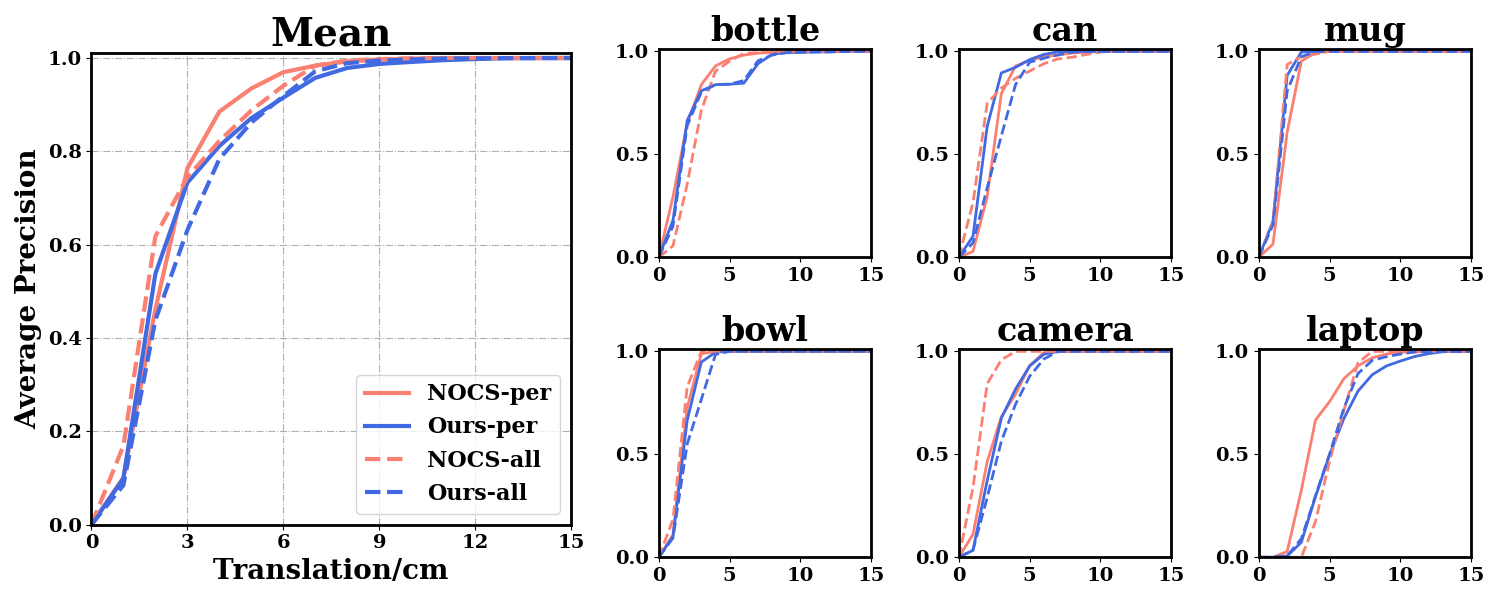}
    \caption{\textbf{Translation}:AP at different translation error thresholds}
 \end{subfigure}
 
 \vfill
 \begin{subfigure}[b]{0.95\textwidth}
    \centering
    \includegraphics[width=1.\linewidth]{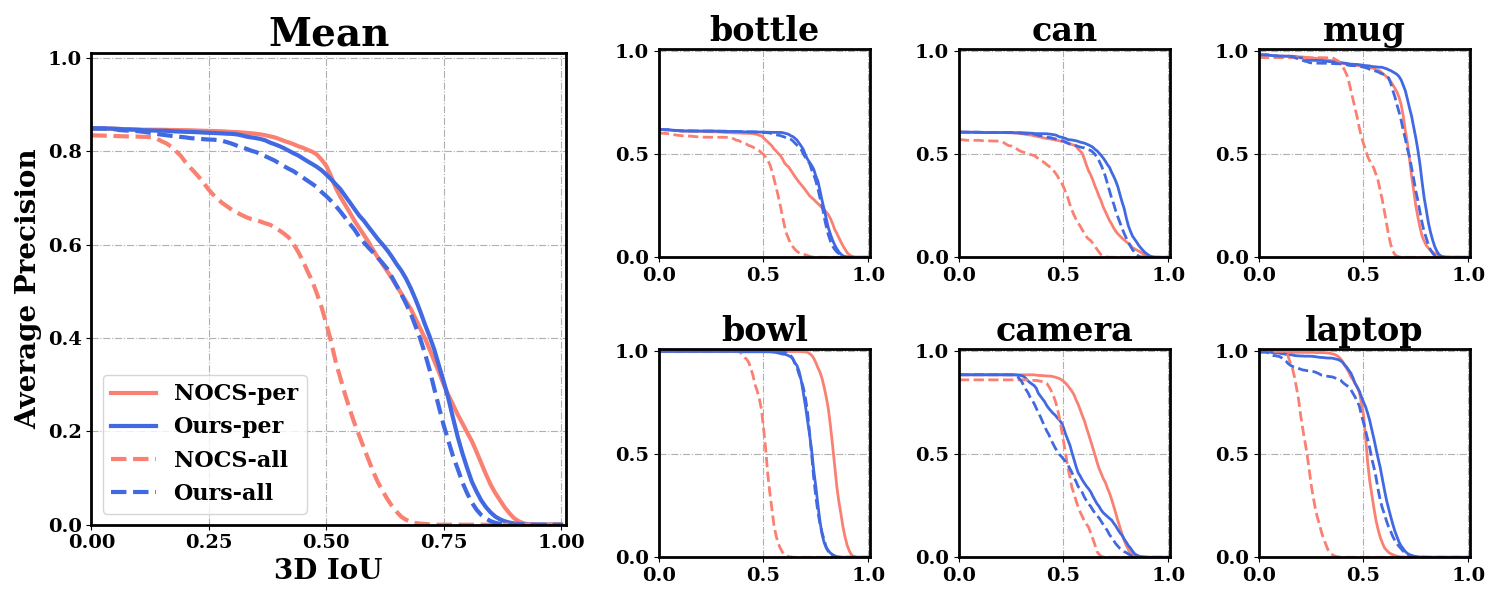}
    \caption{\textbf{3D IoU}:AP at different 3D IoU thresholds}
    \label{fig:real275_iou_nocsall}
 \end{subfigure}
    \caption{Comparison on REAL275 for ours and NOCS~\cite{NOCS2019} regarding setting I (\textit{i.e., Ours-per, NOCS-per}) and setting III (\textit{i.e., Ours-all, NOCS-all}). Note that \textit{NOCS-per} is the original NOCS~\cite{NOCS2019} that trains respective NOCS map branches for different categories. Scope of compared methods is listed in Tab.~\ref{table:nocs_all_setup}. Reported are the average precision under different rotation or translation errors and 3D IoU thresholds. } \label{fig:real275_nocsall}
\end{figure*}

\noindent\textbf{3D IoU} To supplement Sec.~\ref{sec:comparison_unknown} of our main text, we report in Fig.~\ref{fig:real275_iou_nocsall} the 3D IoU for \textit{Ours-per} and compare it with NOCS\cite{NOCS2019} (denoted as \textit{NOCS-per} in Fig.~\ref{fig:real275_iou_nocsall}). 3D IoU not only evaluates the 6D pose estimation but also takes scale estimation into consideration. We note that our closest prior work \cite{Chen2020ECCV} neither reports 3D IoU nor discusses on scale estimation. 

With a simple mean depth comparison for translation and bounding box size comparison for scale estimation, we observe a comparable average performance between \textit{Ours-per} and NOCS~\cite{NOCS2019} with the 3D IoU metric, while our contribution to rotation estimation is clearly supported by the numerical results on rotation error (Fig.~\ref{fig:real275_rot_nocsall}), as is also discussed in the main text.

\begin{figure*}[p]
 \centering
 \begin{subfigure}[b]{0.95\textwidth}
    \centering
    \includegraphics[width=1.\linewidth]{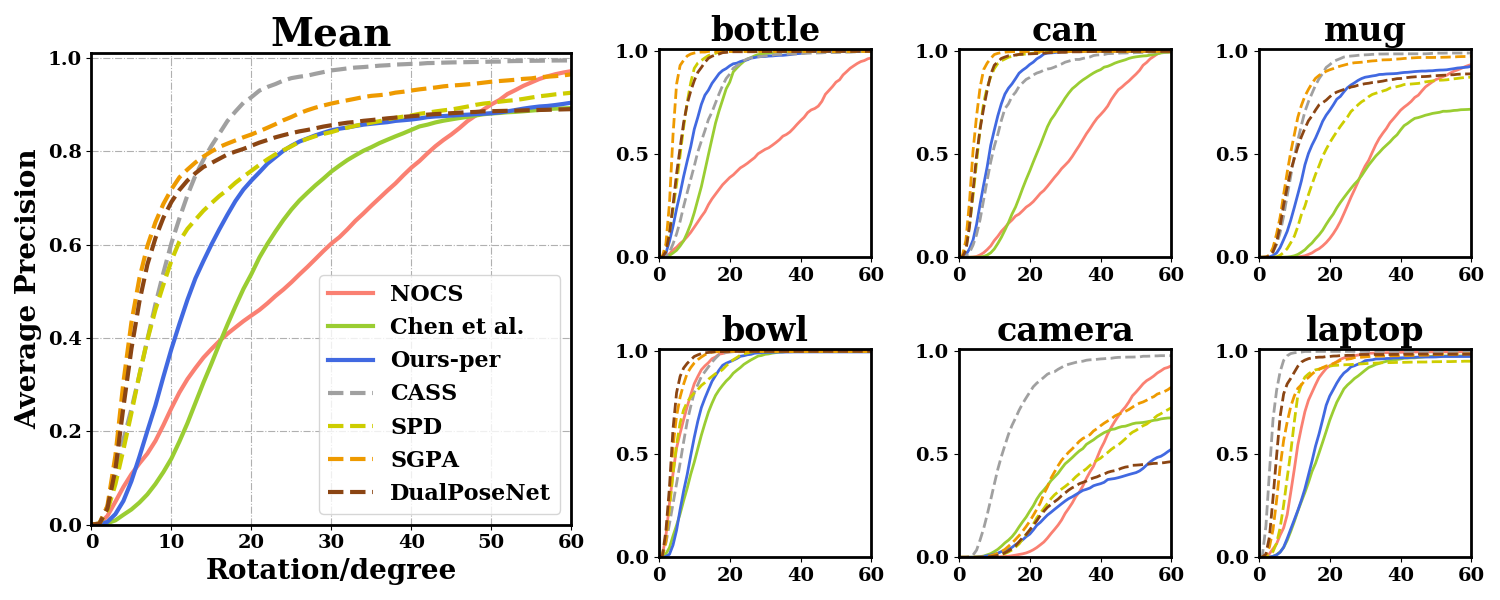}
    \caption{\textbf{Rotation}: AP at different rotation error thresholds}
    \label{fig:real275_rot_supp}
 \end{subfigure}
 \vfill
 \begin{subfigure}[b]{0.95\textwidth}
    \centering
    \includegraphics[width=1.\linewidth]{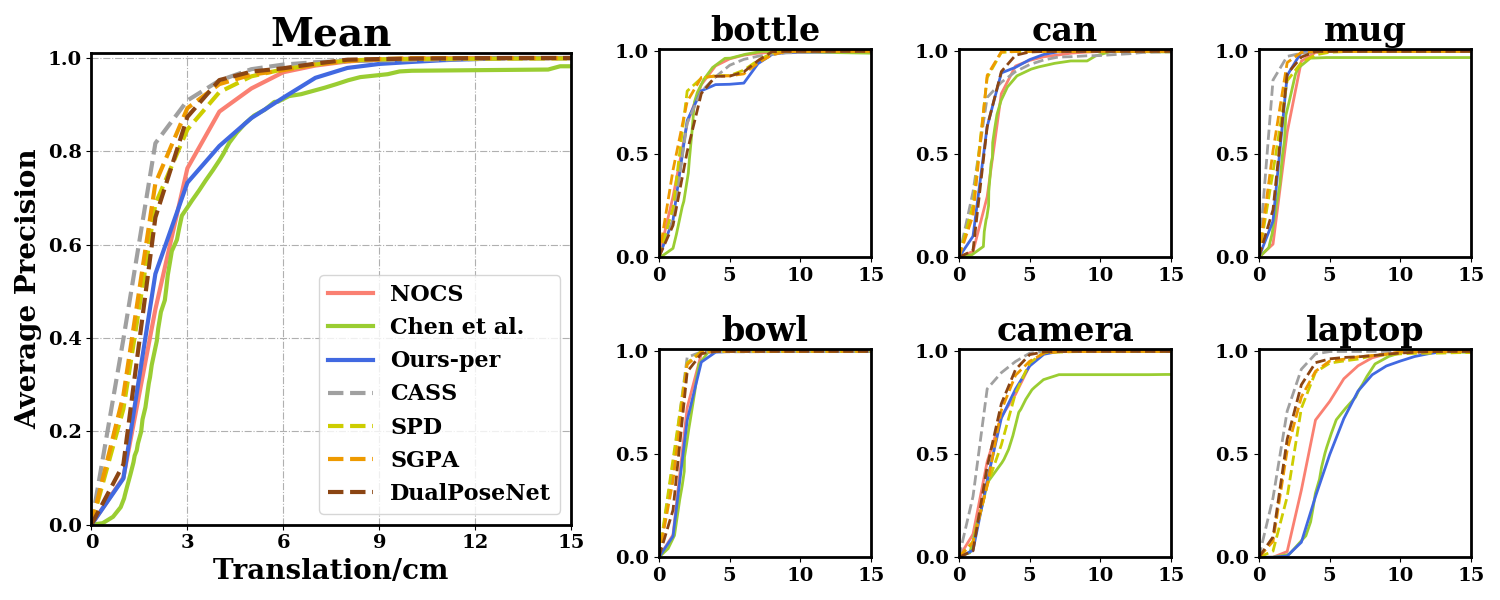}
    \caption{\textbf{Translation}:AP at different translation error thresholds}
    \label{fig:real275_tra_supp}
 \end{subfigure}
 
 \vfill
 \begin{subfigure}[b]{0.95\textwidth}
    \centering
    \includegraphics[width=1.\linewidth]{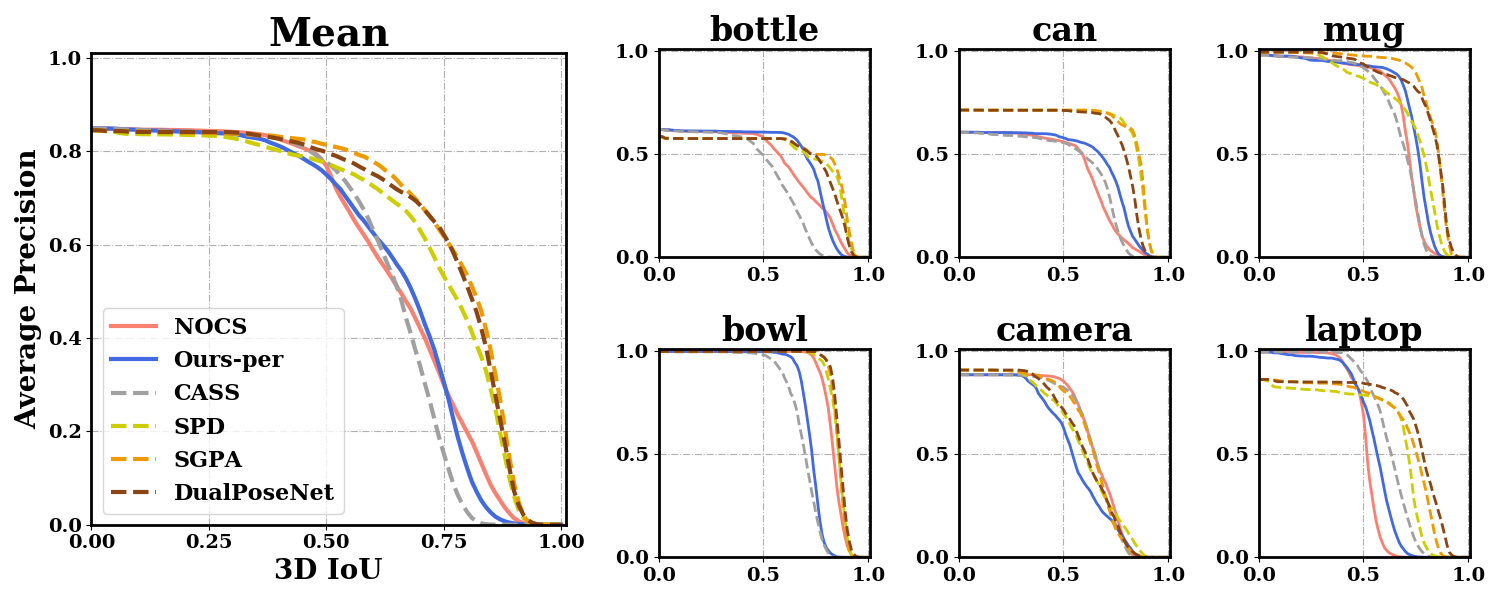}
    \caption{\textbf{3D IoU}:AP at different 3D IoU thresholds}
    \label{fig:real275_iou_supp}
 \end{subfigure}
 \caption{Comparison on REAL275 regarding setting I. Scope of compared methods is listed in Tab.~\ref{table:nocs_setup_supp}. Reported are the average precision under different rotation or translation error and 3D IoU thresholds. CASS~\cite{Chen_2020_CVPR}, SPD~\cite{Tian2020ECCV}, SGPA~\cite{chen2021sgpa} and DualPoseNet~\cite{lin2021dualposenet} fuse RGBD input by networks and train with real data.
 \label{fig:real275_ap_supp}}
\end{figure*}

\noindent\textbf{Comparison with RGBD Fusion Networks}
To provide a more complete view of different category-level pose estimation methods, we include for comparison more methods that fuse RGBD input by networks \cite{Tian2020ECCV,Chen_2020_CVPR,chen2021sgpa,chen2021fs,lin2021dualposenet}, particularly the state-of-the-art CASS\cite{Chen_2020_CVPR}, SPD \cite{Tian2020ECCV}, SGPA\cite{chen2021sgpa} and DualPoseNet \cite{lin2021dualposenet} which provide open-source codes. 
Note that all these methods train pose estimation networks with real data from REAL275\cite{NOCS2019}, which further helps bridge the domain gap between training and testing data. 
The settings and scopes of different methods are listed in Tab.~\ref{table:nocs_setup_supp}. 
Fig.~\ref{fig:real275_ap_supp} shows the mAP of rotation error, translation error and 3D IoU at different thresholds.
As shown here, we note that fusing RGB and depth map with a powerful 3D point cloud processing module can significantly boost pose estimation performance, which can be an important augmentation to our simple but scalable pose estimation framework.

\begin{table*}[tb]
\caption{Comparing the scopes of different methods on REAL275. All methods follow setting I to assign each category a specific network branch, and use query depth for translation estimation.}
\label{table:nocs_setup_supp}
\centering
\begin{center}
\small
\resizebox{1.\textwidth}{!}{
\begin{tabular}{cccccccc}
    \hline 
    &CASS\cite{Chen_2020_CVPR}&SPD\cite{Tian2020ECCV}& DualPoseNet\cite{lin2021dualposenet} & SGPA\cite{chen2021sgpa} & NOCS\protect\cite{NOCS2019} &Chen et al.\protect\cite{Chen2020ECCV} &Ours-per  \\
    \hline
RGB-only network input &$\times$& $\times$  & $\times$  & $\times$ & $\checkmark$ & $\checkmark$ & $\checkmark$ \\ 
Synthetic training data only & $\times$ & $\times$  & $\times$  & $\times$ & $\times$ & $\checkmark$ & $\checkmark$\\
RGB-only for rotation estimation & $\times$ & $\times$ & $\times$ & $\times$ & $\times$ & $\checkmark$ & $\checkmark$\\
\hline
\end{tabular}}
\end{center}
\end{table*}

\section{Supplementary for \textit{Ours-all}(Sec.~\ref{sec:ours_all} of main text, Setting III)} \label{sec:supp_oursall}

\begin{figure*}[p]
 \centering
    \includegraphics[width=1.\linewidth]{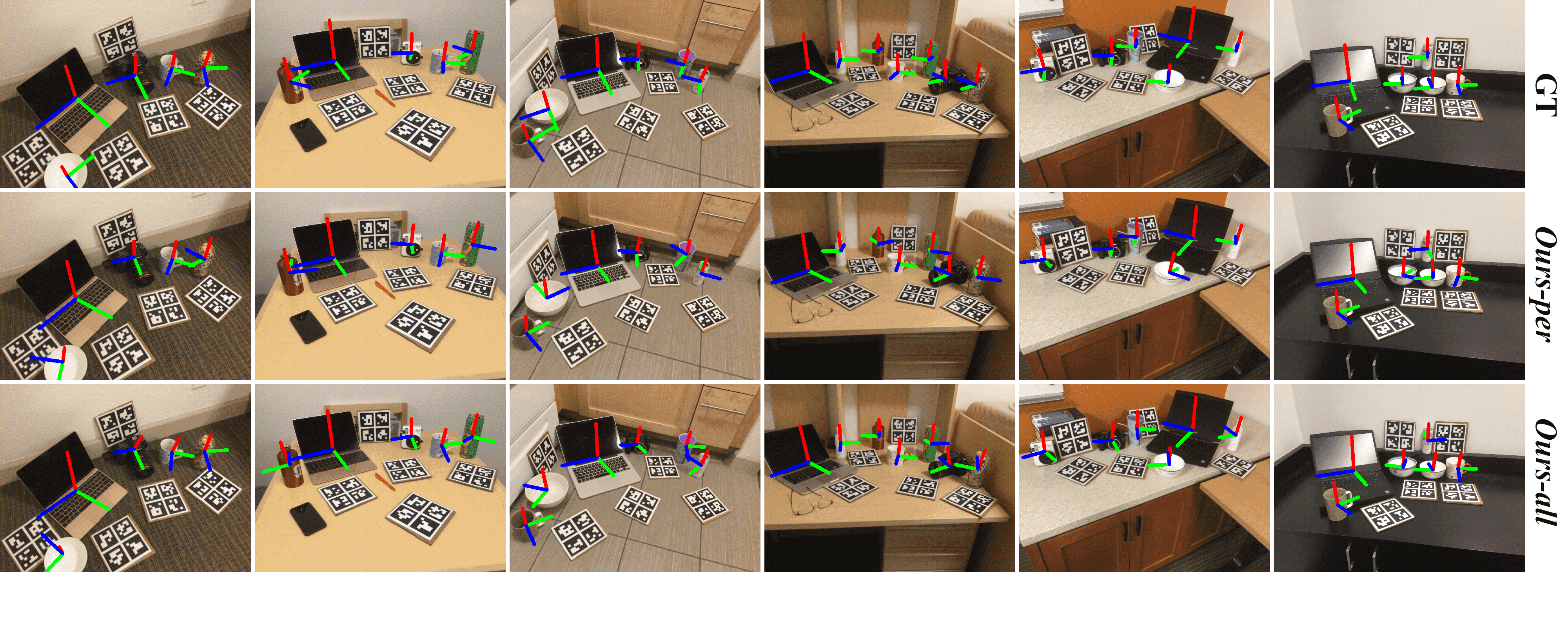}
    \vfill
    \includegraphics[width=1.\linewidth]{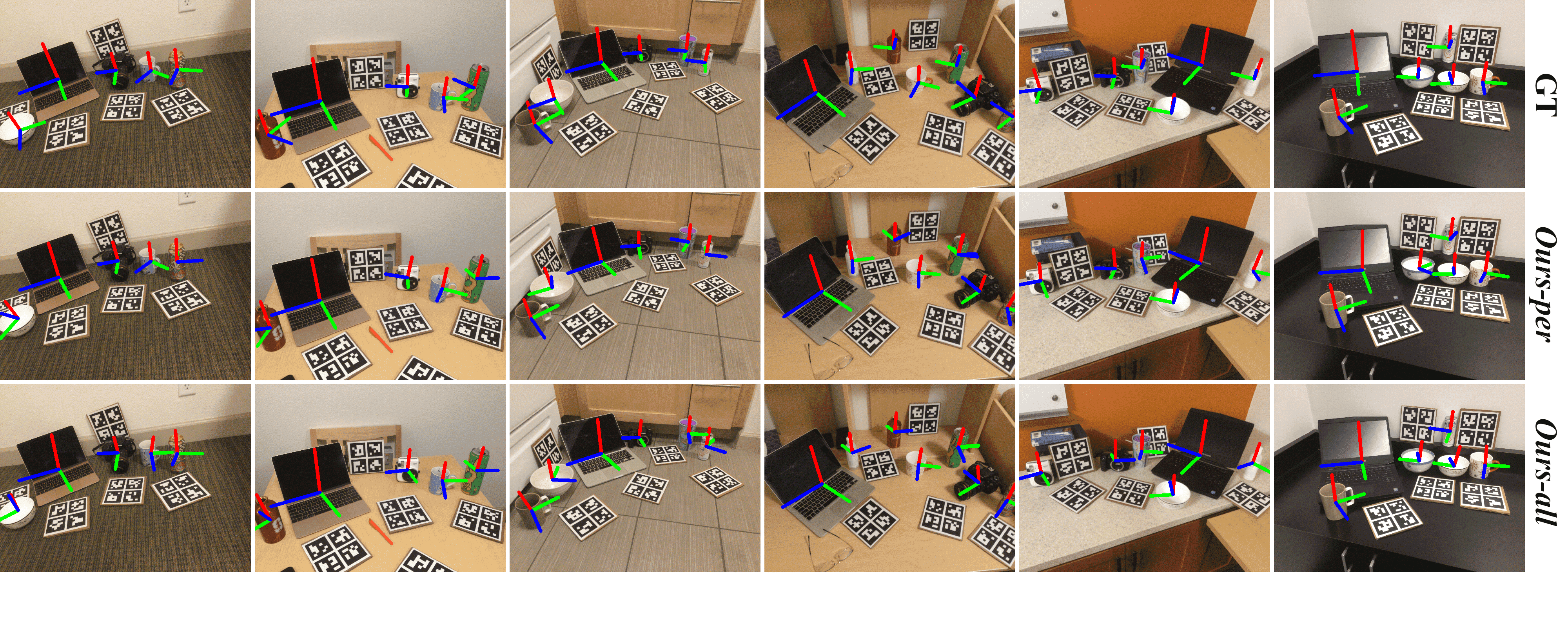}
    \vfill
    \includegraphics[width=1.\linewidth]{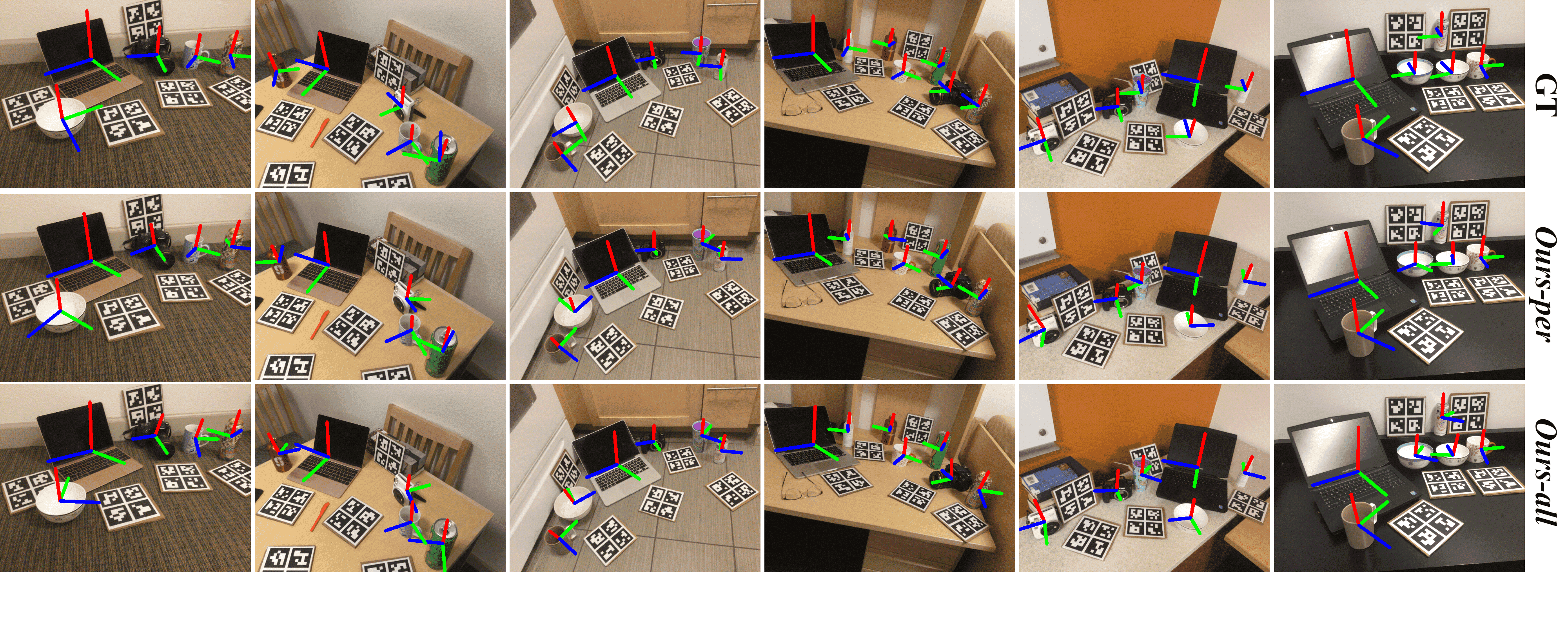}
 
 \caption{Qualitative results on the 6 test scenes of REAL275 for \textit{Ours-per}(Setting I) and \textit{Ours-all}(Setting III). Images of a column belong to a scene.}
 \label{fig:supp_real275}
\end{figure*}

\noindent\textbf{Qualitative Results} 
We provide in Fig.~\ref{fig:supp_real275} qualitative cases of our pose estimation result for \textit{Ours-all} on all 6 scenes of the REAL275 testing set.

\noindent\textbf{Comparison between \textit{Ours-all} and \textit{NOCS-all}}
To supplement the discussion of \textit{Ours-all}, we train \textit{NOCS-all} for NOCS\cite{NOCS2019} by using a common NOCS map head for all 6 categories, where we follow the training configuration of NOCS\cite{NOCS2019} to train on both synthetic data and real data, and adopt the same loss function from NOCS\cite{NOCS2019} by referring to the category label and processing different object symmetry among different categories. We note that with the design of object-conditioned pose code generalization, \textit{Ours-all} exempts the respective loss designs for different categories and adaptively accommodates various object symmetries. We recap the key differences between \textit{Ours-all}, \textit{NOCS-all} in  Tab.~\ref{table:nocs_all_setup}, where we include also the original NOCS\cite{NOCS2019}(denoted as \textit{NOCS-per}) and \textit{Ours-per} for comparison.

\begin{table}[tb]
\centering
\begin{center}
\small
\caption{Comparing the scopes between ours and NOCS regarding setting III (\textit{i.e., Ours-all, NOCS-all}) and setting I (\textit{i.e., Ours-per, NOCS-per}). Note that \textit{NOCS-per} is the original NOCS~\cite{NOCS2019} that trains respective NOCS map branches for different categories. All methods refer to query depth for translation estimation.}
\label{table:nocs_all_setup}
\resizebox{1.\textwidth}{!}{
\begin{tabular}{ccccc}
    \hline 
    &NOCS-per\cite{NOCS2019}&Ours-per &NOCS-all &Ours-all \\
    \hline
Synthetic training data only &$\times$  & $\checkmark$& $\times$  & $\checkmark$ \\
RGB-only for rotation estimation &$\times$ & $\checkmark$ & $\times$ & $\checkmark$\\
Uniform loss for categories with different object symmetry &$\times$ & $\checkmark$& $\times$ & $\checkmark$ \\
Extention to cross-category & $\times$ &$\times$ &$\checkmark$ &$\checkmark$ \\
\hline
\end{tabular}}
\end{center}
\end{table}

We report the mAP for rotation error, translation error and 3D IoU for pose estimation in Fig.~\ref{fig:real275_nocsall}. By comparing the two cross-category variants, we observe that \textit{Ours-all} significantly outperforms \textit{NOCS-all} on rotation estimation for all 6 categories, and reports better performance than \textit{NOCS-all} regarding the mean 3D IoU. Since compared with \textit{NOCS-per},  \textit{NOCS-all} uses a shared NOCS map prediction head for different categories, we hypothesize that the inter-categorical shape variances pose significant challenges for the shared NOCS map branch to exploit shape similarity and extract shape consistency across categories for reliable NOCS map prediction. 
In contrast, \textit{Ours-all} shows competitive performance as discussed in Sec.~\ref{sec:ours_all} of the main text, which is enabled by the shape space metric learning.
Indeed, to address the scalability issue our shape-space contrastive metric learns to model the shape similarity based on instance-level shape discrimination, which allows adaptive exploitation for both inter- and intra-categorical shape features without referring to categorical labels.

\noindent\textbf{Comparison between \textit{Ours-all} and PoseContrast }
We compare with a state-of-the-art PoseContrast\cite{xiao2021posecontrast} that works on the cross-category setting for resolving only the 3D rotation estimation, where we retrain PoseContrast\cite{xiao2021posecontrast} on our setting III of cross-category objects, and evaluate only the rotation accuracy on REAL275 images with GT 2D mask. Results in Table~\ref{table:posecontrast_nocs} demonstrate our superiority in rotation learning for accommodating categories with different object symmetry, where PoseContrast\cite{xiao2021posecontrast} does not handle objects with different symmetries as well as we do.

\begin{table}[tb]
\centering
\begin{center}
\small
\caption{Comparison of rotation estimation with PoseContrast\cite{xiao2021posecontrast} on REAL275 with GT 2D mask. We report rotation accuracy under the error threshold of $30^{\circ}$. }
\label{table:posecontrast_nocs}
\resizebox{.8\textwidth}{!}{
\begin{tabular}{c|c|c|c|c|c|c|c}
\hline
    &Bottle&Bowl&Camera&Can&Laptop&Mug&Ave.\\
\hline
PoseContrast\cite{xiao2021posecontrast}&83.5&81.9&11.7&87.2&30.0&35.8&55.0\\
\textit{Ours-all}&\textbf{96.8}&\textbf{99.8}&\textbf{58.6}&\textbf{99.3}&\textbf{93.9}&\textbf{91.0}&\textbf{89.9}\\
\hline
\end{tabular}}
\end{center}
\end{table}

\begin{figure}[tb]
 \centering
 \includegraphics[width=0.8\textwidth]{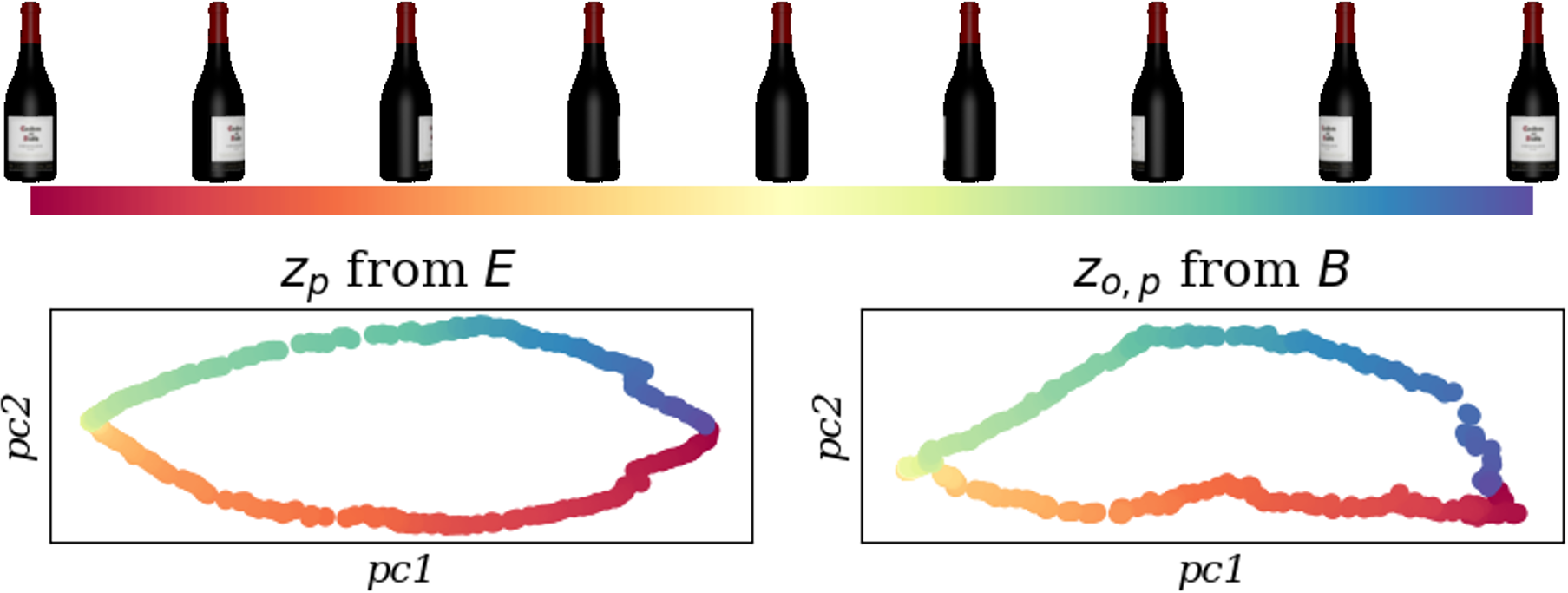}
    \caption{Visualization of pose codes for a textured wine bottle viewed in $360^\circ$ as it rotates axially (upper). We show the top two PCA projection of pose codes $\vb*{z}_p, \vb*{z}_{o,p}$ produced by \textit{Ours-all} network (bottom), with point color encoding the viewpoints above. Texture has enabled distinguishing axial symmetries.}
    \label{fig:supp_symmetry_with_texture}
\end{figure}

\noindent\textbf{Visualization of Pose Codes for the Textured Bottle}
For symmetric objects with textural features solving the pose ambiguity, we note that our network can indeed tell the different poses by referring to the textures. Taking a textured wine bottle from the CAMERA training objects as an example, we rotate it around its symmetry axis and respectively inspect the top 2 PCA projections for pose codes $\vb*{z}_p, \vb*{z}_{o,p}$. The visualization result is in  Fig.~\ref{fig:supp_symmetry_with_texture}, where the pose codes well describe the different textural appearance.

\section{Results on T-LESS (Sec.~\ref{sec:comparison_known} of main text, Setting II) }\label{sec:supp_tless}

\begin{table*}[tb]
\caption{Object recall rate with $e_{VSD}<0.3$ for our full 2D detection+pose estimation pipeline, where we train on Obj. 1-18 and report on all 30 objects. Instances with visible portion $>10\%$ for all T-LESS Primesense images are considered.}\label{table:tless_detper}
\begin{subtable}[t]{1.0\textwidth}
\begin{center}{
\begin{tabular}{c p{0.8cm}<{\centering}p{0.8cm}<{\centering}p{0.8cm}<{\centering}p{0.8cm}<{\centering}p{0.8cm}<{\centering}p{0.8cm}<{\centering}p{0.8cm}<{\centering}p{0.8cm}<{\centering}p{0.8cm}<{\centering}p{0.8cm}<{\centering}p{0.8cm}<{\centering}p{0.8cm}<{\centering}}
\hline
Obj-id&1&2&3&4&5&6&7&8&9&10\\
\hline
Recall rate&26.05&16.85&33.52&25.43&51.73&47.90&19.53&21.85&32.88&44.50\\
\hline
\end{tabular}
}
\end{center}
\end{subtable}
\begin{subtable}[t]{1.0\textwidth}
\begin{center}{
\begin{tabular}{c p{0.8cm}<{\centering}p{0.8cm}<{\centering}p{0.8cm}<{\centering}p{0.8cm}<{\centering}p{0.8cm}<{\centering}p{0.8cm}<{\centering}p{0.8cm}<{\centering}p{0.8cm}<{\centering}p{0.8cm}<{\centering}p{0.8cm}<{\centering}p{0.8cm}<{\centering}p{0.8cm}<{\centering}}
\hline
Obj-id&11&12&13&14&15&16&17&18&19&20\\
\hline
Recall rate&21.14&42.97&41.44&35.28&42.77&41.23&49.06&65.90&22.77&24.09\\
\hline
\end{tabular}
}
\end{center}
\end{subtable}
\begin{subtable}[t]{1.0\textwidth}
\begin{center}{
\begin{tabular}{c p{0.8cm}<{\centering}p{0.8cm}<{\centering}p{0.8cm}<{\centering}p{0.8cm}<{\centering}p{0.8cm}<{\centering}p{0.8cm}<{\centering}p{0.8cm}<{\centering}p{0.8cm}<{\centering}p{0.8cm}<{\centering}p{0.8cm}<{\centering}p{0.8cm}<{\centering}p{0.8cm}<{\centering}}
\hline
Obj-id&21&22&23&24&25&26&27&28&29&30\\
\hline
Recall rate&33.22&20.44&18.18&42.32&34.28&45.73&27.42&43.67&52.91&35.66\\
\hline
\end{tabular}
}
\end{center}
\end{subtable}
\end{table*}

\begin{figure*}[tb]
 \centering
 \includegraphics[width=0.85\textwidth]{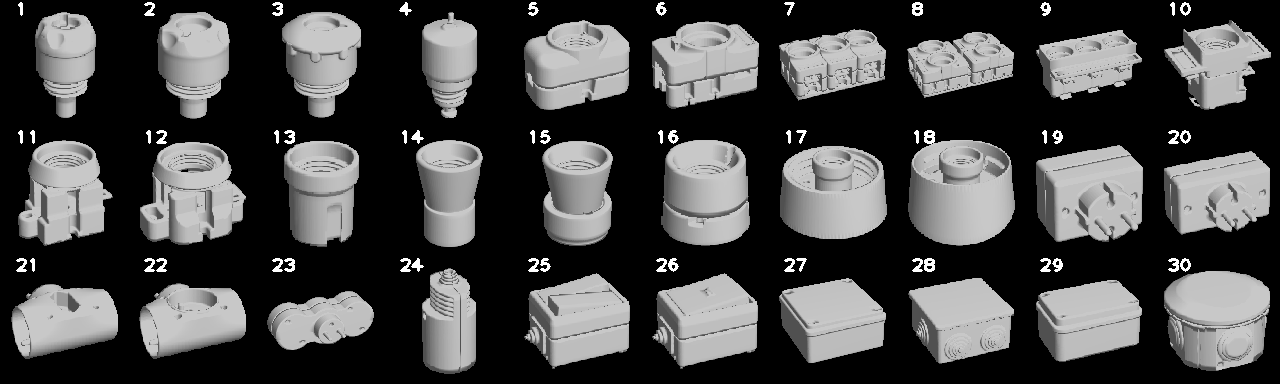}
 \caption{30 objects included in the T-LESS dataset \cite{hodan2017tless}.}
 \label{fig:tless_meshes}
\end{figure*}

\noindent\textbf{30 Objects Included in T-LESS} 
In Fig.~\ref{fig:tless_meshes}, we visualize the CAD models of all 30 objects in T-LESS\cite{hodan2017tless}. 
As we train on Obj. 1-18 while leaving Obj. 19-30 as novel objects in the testing stage, the drastic shape variance and different rotational symmetry can be observed when comparing among the training objects, and comparing between the training objects and the unseen test objects. 
Our method accommodates these different shapes by a single network.

\noindent\textbf{Per-Object Recall Rate for the Full ``Detection+Pose Estimation'' Pipeline. }
For the pose estimation results with 2D bounding boxes detected by MaskR-CNN \cite{he2017mask} (Tab.~\ref{table:tless_2D} in the main text), we report the detailed recall rate with $e_{VSD}<0.3$ for each of the 30 objects in Tab.~\ref{table:tless_detper}. 
Here we have followed the single object single instance protocol as described in \cite{hodan2018bop}. 

\begin{figure*}[p]
 \centering
 \includegraphics[width=0.99\textwidth]{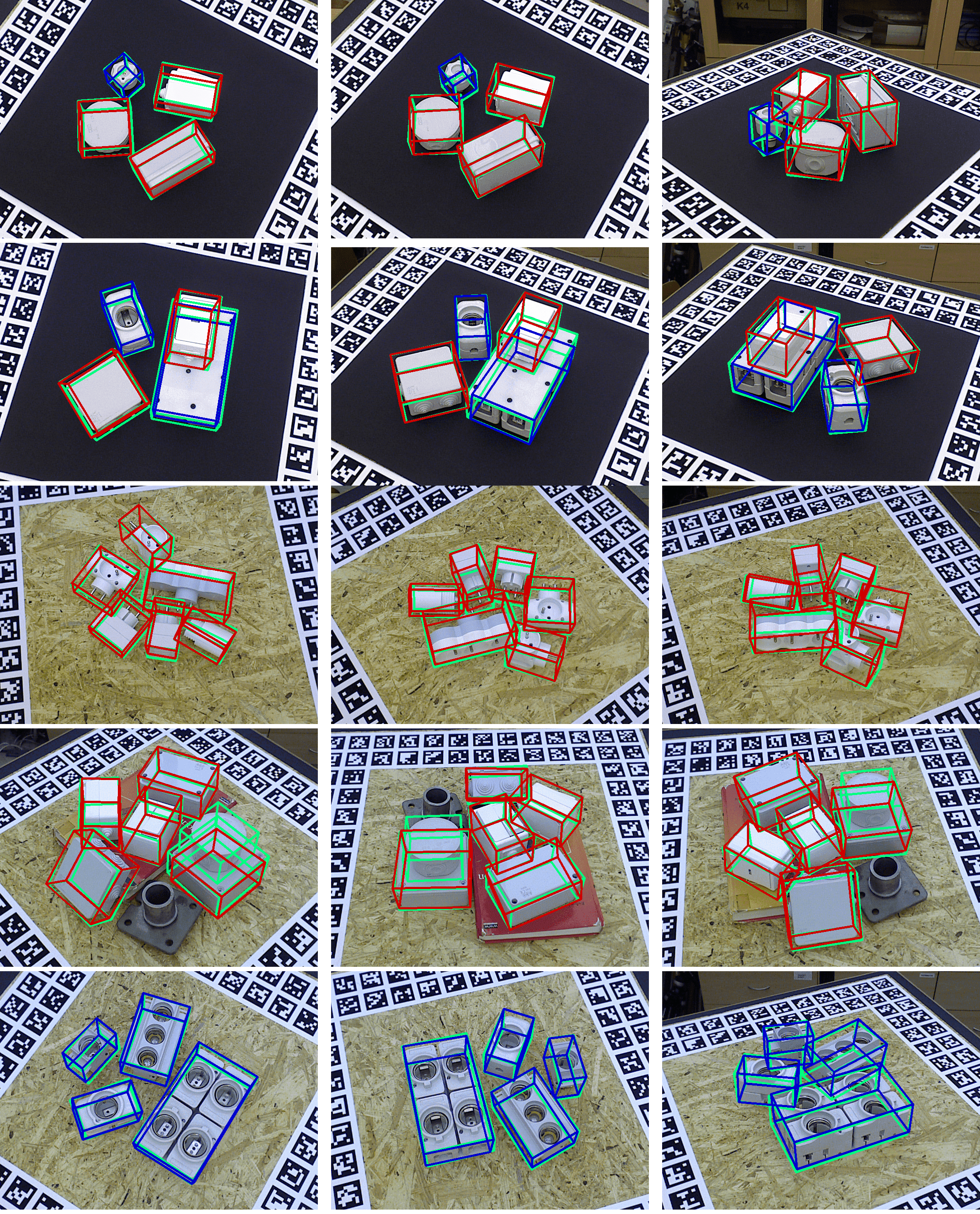}
 \caption{Qualitative cases on T-LESS of setting II. Blue and red boxes respectively indicate our estimation on trained objects and unseen objects, with green box indicating the groundtruth.}
 \label{fig:supp_tless}
\end{figure*}

\noindent\textbf{Qualitative Results for the Full ``Detection+Pose Estimation'' Pipeline. }We provide in Fig.~\ref{fig:supp_tless} more qualitative cases of our  full ``detection+pose estimation'' pipeline. We notice false negatives caused by failures of 2D detection. 
However, for the detected instances, our pose estimation could well process both trained and novel objects, even under the challenging conditions of cluttering and partial occlusion.

\section{Instance-Level Pose Estimation}\label{sec:supp_instlevel}

\begin{table*}[tb]
\caption{Comparison of pose estimation on T-LESS dataset, following the ViVo setting of BOP Challenge 2020\cite{hodan2020bop}. All reported methods test with single RGB image. $\dagger$ indicates having post-refinement. Bold statistics in black, red, and blue respectively indicate the best, second best, and third best.}
\label{table:tless_instance}
\centering
\begin{center}
\small
\resizebox{0.85\textwidth}{!}{
\begin{tabular}{ccccccc}
    \hline
    \textbf{Method} &  \textbf{Training Image Type} & \textbf{$AR_{VSD}$} & \textbf{$AR_{MSSD}$} & \textbf{$AR_{MSPD}$} & \textbf{$AR$} & \textbf{Time(s)}\\
    \hline
    AAE\cite{sundermeyer2019augmented} & Real+Syn & 0.196 & 0.211 & 0.504 & 0.304 & \secb{0.194}\\
    Pix2Pose\cite{park2019pix2pose} & Real & 0.261 & 0.296 & 0.476 & 0.344 & 1.084\\
    CDPN\cite{li2019cdpn} & Syn & 0.303 & \thrb{0.338} & 0.579 & 0.407 & 1.849 \\
    EPOS\cite{hodan2020epos} & Syn & \secb{0.380} & \secb{0.403} & \thrb{0.619} & \secb{0.467} & 1.992 \\
    CosyPose\cite{labbe2020cosypose}${\dagger}$ & Syn & \textbf{0.571} & \textbf{0.589} & \textbf{0.761} & \textbf{0.640} & \thrb{0.493} \\
    Ours                        & Syn & \thrb{0.316} & {0.326} & \secb{0.650} & \thrb{0.431} & \textbf{0.118}\\
    \hline
\end{tabular}}
\end{center}
\end{table*}

To explore the limiting case of instance level pose estimation where all objects are used for training, we compare with the state-of-the-art instance-level methods \cite{sundermeyer2018implicit,sundermeyer2019augmented,park2019pix2pose,li2019cdpn,labbe2020cosypose,hodan2020epos} on the T-LESS dataset, where we follow the setting of BOP Challenge 2020\cite{hodan2020bop} to evaluate our method on the ViVo task (varying number of instances of a varying number of objects) for 6D localization.

\noindent\textbf{Metrics} Three pose-error metrics are measured in the BOP challenge\cite{hodan2020bop}: Visible Surface Discrepancy (VSD), Maximum Symmetry-Aware Surface Distance (MSSD), and Maximum Symmetry-Aware Projection Distance (MSPD). 
These metrics are invariant under symmetry ambiguity.

For each metric, we follow the BOP Challenge\cite{hodan2020bop} to calculate the average recall rate under a list of thresholds of correctness (denoted $AR_{VSD}$, $AR_{MSSD}$, $AR_{MSPD}$), as well as the overall average recall $AR=(AR_{VSD}+AR_{MSSD}+AR_{MSPD})/3$. 

\noindent\textbf{Training Strategy} We train our network on all 30 T-LESS models. Our training set combines synthetic images with 92232 poses per object rendered by the pipeline described in Sec.\ref{sec:training_data}, and the photorealistic training images from the BOP challenge, which are generated by a physically-based renderer (PBR)\cite{denninger2019blenderproc}. This combination helps us to learn regular latent spaces and to better bridge the synthetic-to-real domain gap. Our 2D detector is the MaskR-CNN adopted from CosyPose \cite{labbe2020cosypose}.

\noindent\textbf{Results and Comparison} We test our method on a machine with i7-6700K 4GHz CPU and Nvidia GTX 1080 GPU, and report the performance and the average running time per image in Tab.~\ref{table:tless_instance}, where we compare with other single RGB-based methods from the BOP leaderboard. 
Note that CDPN\cite{li2019cdpn}, EPOS\cite{hodan2020epos} and CosyPose\cite{labbe2020cosypose} are trained with synthetic PBR images, while Pix2Pose\cite{park2019pix2pose} is trained with real images.

Among the methods listed in Tab.~\ref{table:tless_instance}, ours is capable of providing a fast yet reliable pose estimation, which could serve as initialization and be further refined if applicable. 
Specifically, we rank third by the overall $AR$, and second by $AR_{MSPD}$ which evaluates the 2D projections and thus exempts from the influence of inaccurate depth estimation. 
Meanwhile, building on an autoencoding pipeline (same as AAE \cite{sundermeyer2019augmented}) enables our approach to do fast pose estimation for detected instances, with lower time cost compared with the other methods \cite{park2019pix2pose,li2019cdpn,hodan2020epos,labbe2020cosypose}. Cosypose \cite{labbe2020cosypose} has the best performance among all error metrics; however, it relies on post-refinement with a regression network after the initial pose estimation. 
EPOS \cite{hodan2020epos} also outperforms ours overall, as it uses PnP-RANSAC on many-to-many 2D-3D correspondences for reliable rotation and translation with a significant time cost. 
In comparison, our estimated depth from the pinhole camera model is sensitive to the inaccurate size of the detected 2D bounding box, and we show better performance on $AR_{MSPD}$ where the influence of depth error is minimized.

\clearpage

\end{document}